%% file: main.tex
\documentclass[runningheads]{llncs}

\PassOptionsToPackage{table}{xcolor}
\usepackage{eccv}

\definecolor{eccvblue}{rgb}{0.21,0.49,0.74}
\usepackage{eccvabbrv}

\usepackage[utf8]{inputenc}
\usepackage[T1]{fontenc}
\usepackage{bm,xspace,comment,multirow,balance,url,booktabs}
\usepackage{etoolbox,siunitx,calc,pifont,hologo,adjustbox,enumitem}
\usepackage[normalem]{ulem}
\usepackage{amsfonts,nicefrac,microtype,wrapfig,graphicx}
\usepackage{caption,subcaption}

\usepackage{amsmath,amssymb,mathtools,multicol,makecell,array}
\usepackage{float}
\usepackage{algorithm,algorithmicx,algpseudocode}
\newcommand{\cmark}{\textcolor{green}{\ding{51}}}%
\newcommand{\xmark}{\textcolor{red}{\ding{55}}}%
\newcommand{\dash}{\textcolor{orange}{\rule[0.5ex]{0.9em}{2pt}}}%

\usepackage[breaklinks,colorlinks,urlcolor=black]{hyperref}

\newcommand{\ourmethod}{\textsc{UniQueR}\xspace}

\begin{document}

\title{\ourmethod: Unified Query-based Feedforward 3D Reconstruction} 

\titlerunning{\ourmethod}

\author{Chensheng Peng\inst{1,2}\textsuperscript{*} \and
Quentin Herau\inst{1} \and
Jiezhi Yang\inst{1} \and
Yichen Xie\inst{1,2}\textsuperscript{*} \and
Yihan Hu\inst{1} \and
Wenzhao Zheng\inst{2} \and
Matthew Strong\inst{1,3}\textsuperscript{*} \and
Masayoshi Tomizuka\inst{2} \and
Wei Zhan\inst{1,2}\textsuperscript{\textdagger}}

\authorrunning{C.~Peng et al.}

\institute{
\textsuperscript{1}Applied Intuition
\quad
\textsuperscript{2}UC Berkeley
\quad
\textsuperscript{3}Stanford University
}

\maketitle
\begingroup
\renewcommand{\thefootnote}{}
\footnotetext{{Project page: \url{https://uniquer3d.github.io/}}}
\endgroup
\begingroup
\renewcommand{\thefootnote}{\fnsymbol{footnote}}
\footnotetext[1]{Work done during an internship at Applied Intuition.}
\footnotetext[2]{Corresponding author: \email{wei.zhan@applied.co}.}
\endgroup

\input{sec/0_abstract}

\input{sec/1_intro}

\input{sec/2_related_work}

\input{sec/3_method}

\input{sec/4_experiments}

\input{sec/5_conclusion}

\bibliographystyle{splncs04}
\bibliography{main}
\newpage
\input{sec/X_supplement}

\end{document}

%% file: sec/0_abstract.tex
\begin{abstract}
We present \ourmethod, a unified query-based feedforward framework for efficient and accurate 3D reconstruction from unposed images. Existing feedforward models such as DUSt3R, VGGT, and AnySplat typically predict per-pixel point maps or pixel-aligned Gaussians, which remain fundamentally 2.5D and limited to visible surfaces. In contrast, \ourmethod formulates reconstruction as a sparse 3D query inference problem. Our model learns a compact set of 3D anchor points that act as explicit geometric queries, enabling the network to infer scene structure---including geometry in occluded regions---in a single forward pass. Each query encodes spatial and appearance priors directly in global 3D space (instead of per-frame camera space) and spawns a set of 3D Gaussians for differentiable rendering. By leveraging unified query interactions across multi-view features and a decoupled cross-attention design, \ourmethod achieves strong geometric expressiveness while substantially reducing memory and computational cost. Experiments on Mip-NeRF 360 and VR-NeRF demonstrate that \ourmethod surpasses recent feedforward baselines in both rendering quality and geometric accuracy, using an order of magnitude fewer primitives than dense alternatives. 
\keywords{Feedforward reconstruction \and Gaussian splatting \and Queries}
\end{abstract}

%% file: sec/1_intro.tex
\section{Introduction}
\label{sec:intro}

\begin{figure*}[t]
    \centering
    \includegraphics[width=\linewidth]{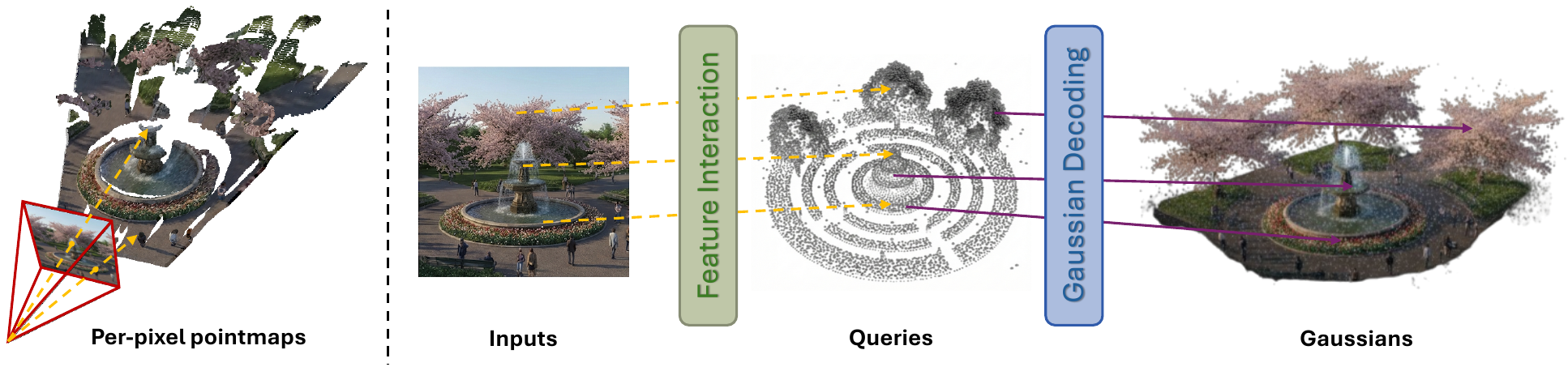}
    \caption{{\textbf{Illustration of image-aligned and query-based pipelines.} Given input images, \ourmethod predicts a sparse set of 3D queries that spawn Gaussians in a shared scene coordinate system. The query-based prediction can allocate primitives beyond observed input surfaces.  Unlike pixel-aligned methods (e.g., AnySplat) that produce holes in unobserved areas, our query-based representation enables more complete 3D reconstruction with accurate geometry.}}
    \label{fig:teaser}
\end{figure*}

3D reconstruction from 2D captures is a fundamental task in computer vision \cite{wang2025vggt}, with wide-ranging applications in robotics, autonomous driving, and digital content creation. The goal is to recover accurate 3D geometry and appearance from one or more 2D observations, enabling machines to perceive and interact with the world in a spatially coherent manner.

Traditional methods such as Structure-from-Motion (SfM) \cite{schonberger2016colmap}, Multi-View Stereo (MVS) \cite{yao2018mvsnet}, and Simultaneous Localization and Mapping (SLAM) \cite{bloesch2018codeslam} rely on geometric optimization and feature matching to estimate camera poses and dense point clouds. While effective under controlled settings, they often struggle with visual ambiguity, sparse viewpoints, or textureless regions.

With the rise of deep learning, optimization-based representations---notably Neural Radiance Fields (NeRFs) \cite{mildenhall2020nerf} and 3D Gaussian Splatting~\cite{kerbl2023gaussian}---have produced impressive visual fidelity and multi-view consistency. However, their per-scene optimization nature limits scalability, generalization, and runtime efficiency. Such methods cannot leverage large-scale data to learn transferable geometric priors, and reconstruction for a new scene remains computationally expensive.
{Meanwhile, the research community has increasingly explored feedforward 3D reconstruction frameworks that support real-time inference, end-to-end learning, and data-driven geometric priors. Unlike optimization-based methods that require per-scene fitting, these architectures learn a shared reconstruction model across large-scale datasets, making single-pass inference possible on in-the-wild captures.}

Existing works such as Dust3R \cite{wang2024dust3r}, VGGT \cite{wang2025vggt}, and Pi3 \cite{wang2025pi3permutationequivariantvisualgeometry} demonstrate that 3D geometry can be recovered directly from 2D observations in a single forward pass. These models typically extract multi-view features using transformers or cost volumes and predict intermediate 2.5D representations such as depth maps, normal maps, or point maps.
{Large-scale training with depth priors enables these models to transfer learned geometry across scenes.}
This paradigm shift highlights the potential of feedforward architectures to serve as general-purpose 3D perception backbones for robotics and embodied AI.

Building on this idea, recent methods have extended feedforward prediction beyond point-based outputs to more expressive representations. Approaches such as MVSplat \cite{chen2024mvsplat}, {NoPoSplat} \cite{ye2024poseproblemsurprisinglysimple}, FLARE \cite{zhang2025flare} and AnySplat \cite{jiang2025anysplat} generate sets of 3D Gaussians directly from images, representing both geometry and appearance in a compact and differentiable form. These Gaussian-based outputs allow efficient rendering with photorealistic details and natural view-dependent effects, bridging the gap between feedforward prediction and neural representations. Moreover, they unify geometry and appearance modeling under a single representation, enabling downstream tasks like relighting, editing, and novel view synthesis.

However, despite their impressive progress, these feedforward methods still rely on 2.5D representations: they predict surface-aligned Gaussians or camera-view features that only capture the visible surfaces within the observed viewpoints. As a result, they remain inherently view-anchored. Because the learned features are tied to specific camera projections, these models struggle to allocate geometry to occluded or unseen regions, leading to holes and artifacts in novel views that deviate from the input cameras.

To address this limitation, we propose \ourmethod, a Unified Query-based Feedforward 3D Reconstruction framework that directly operates in 3D space. Inspired by query-based architectures from object detection and scene understanding (e.g., DETR \cite{carion2020endtoendobjectdetectiontransformers}, DETR3D \cite{wang2023exploringobjectcentrictemporalmodeling}, PETR \cite{liu2022petrpositionembeddingtransformation}), we introduce learnable 3D queries as the core scene representation. Each query acts as a spatial anchor in 3D, encoding geometric and appearance attributes. Without requiring ground-truth 3D supervision, we leverage Gaussian Splatting~\cite{kerbl2023gaussian} as a differentiable rendering bridge between 3D queries and 2D observations. Through a cross attention mechanism, the learnable 3D queries interact with the multi-view images and integrate features into their own embeddings. Each query then spawns a set of Gaussians, which are rendered to 2D images with color and depth---enabling supervision from RGB images and depth maps.

Crucially, our training procedure supervises the spawned Gaussians using novel views beyond the input cameras. For example, with 2 input views, we would rasterize the inferred Gaussians to 4 novel poses and supervise the renderings with corresponding ground truth images. Such paradigm encourages 3D queries to not only recover the observed surfaces from inputs but also allocate Gaussians to regions not directly observed. Combined with a decoupled cross attention module for efficient feature interaction, \ourmethod can reconstruct more complete scenes even under sparse or partial observations with a faster speed.

Our main contributions are as follows:
\begin{itemize}
\item We introduce \ourmethod, a scene-level feedforward 3D reconstruction framework based on learnable 3D queries that decouple geometry from input viewpoints, enabling the model to place Gaussians in unobserved regions.
\item We devise a decoupled cross-attention mechanism for integrating per-view image features into global learnable queries, which scales efficiently to a large number of input views.
\end{itemize}

We demonstrate comparable novel-view synthesis \emph{and} geometric accuracy on Mip-NeRF 360~\cite{barron2022mipnerf360unboundedantialiased} and VR-NeRF~\cite{Xu_2023}, using an order of magnitude fewer Gaussians and less memory than dense feedforward baselines.

%% file: sec/2_related_work.tex
\section{Related Work}
\label{sec:related_work}

\textbf{Optimization-based 3D reconstruction. }
{Traditional pipelines for 3D reconstruction and novel view synthesis typically follow two sequential stages. The first estimates camera parameters using geometric and multi-view stereo methods such as Structure-from-Motion (SfM)~\cite{crandall2011discrete,wilson2014robust,sweeney2015optimizing,snavely2006photo,agarwal2009building,wu2013towards,schonberger2016colmap} and stereo matching~\cite{furukawa2009furu,schonberger2016pixelwise,galliani2015massively}. SfM systems---most notably COLMAP~\cite{schonberger2016colmap}---align images via feature correspondences to generate sparse point clouds, refined through bundle adjustment~\cite{triggs2000bundle,agarwal2010bundle,wu2011multicore,yu2024sim}. Related approaches such as V-SLAM~\cite{davison2007monoslam,klein2007parallel,newcombe2011dtam} extend these ideas to real-time tracking and mapping.}
Although robust, these pipelines incur high computational cost due to hand-crafted feature extraction, iterative correspondence search, and large-scale nonlinear optimization~\cite{pan2024globalstructurefrommotionrevisited}, leading to significant runtime overhead and limited scalability.

Following geometry-based reconstruction, per-scene optimization has become standard for novel view synthesis. Differentiable rendering methods like NeRF~\cite{mildenhall2020nerf} and its variants~\cite{barron2022mip,kerbl2023gaussian,barron2023zip,wang2021neus,10.1007/978-3-031-73024-5_14,yariv2021volsdf,jang2021codenerf,jang2024nvist,huang20242d,herau2023moisst,herau2024soac,herau20243dgs,herau2025pose,yang2024carff} optimize neural scene representations per scene given known camera poses. While highly accurate, they require minutes to hours of optimization for a single scene.

Recent approaches such as hash-encoded fields (InstantNGP)~\cite{mueller2022instant} and 3D Gaussian Splatting~\cite{kerbl2023gaussian} substantially reduce training time, but still depend on nontrivial per-scene optimization, limiting their practicality for real-time or large-scale reconstruction.

\input{tables/comparison_of_representations}

\vspace{5pt}
\textbf{Feedforward reconstruction.}
By leveraging large-scale 3D datasets, recent feedforward 3D reconstruction methods directly regress pixel-aligned point maps or Gaussian primitives from posed multi-view images. Splatter Image~\cite{szymanowicz2024splatterimageultrafastsingleview} predicts pixel-aligned Gaussians from a single image, while PixelSplat~\cite{charatan2024pixelsplat3dgaussiansplats} extends this formulation to stereo pairs. MVSplat~\cite{chen2024mvsplat} further generalizes to variable-length sequences but still relies on known input poses. In parallel, feedforward models for monocular or multi-view depth estimation~\cite{dexheimer2023learning,yin2023metric3d,ke2024repurposing,yao2018mvsnet,duzceker2021deepvideomvs,sayed2022simplerecon}, optical flow~\cite{teed2020raft}, and point tracking~\cite{doersch2023tapir,karaev2023cotracker,xiao2024spatialtracker} have steadily advanced the ability to learn geometry from raw visual input.
{DUSt3R~\cite{wang2024dust3r} maps image pairs to dense point maps followed by global alignment, while MASt3R~\cite{leroy2024groundingimagematching3d} improves matching through dense local descriptors. Building on these foundations, Splatt3R~\cite{smart2024splatt3rzeroshotgaussiansplatting} and NoPoSplat~\cite{ye2024poseproblemsurprisinglysimple} directly regress Gaussian primitives from uncalibrated image pairs. Spann3R~\cite{wang20243dreconstructionspatialmemory}, MUSt3R~\cite{cabon2025must3r}, and PreF3R~\cite{chen2024pref3rposefreefeedforward3d} incorporate memory or recurrent processing for longer sequences. GS-LRM~\cite{zhang2024gslrm} instead predicts per-pixel Gaussians from posed sparse views and demonstrates both object- and scene-level reconstruction, making it a direct precursor in feedforward Gaussian decoding rather than an exclusively object-centric method.}

{A recent line of work jointly infers camera parameters and 3D structure. VGGT~\cite{wang2025vggt} and AnySplat~\cite{jiang2025anysplat} provide end-to-end pose-free reconstruction and rendering, while MapAnything~\cite{keetha2025mapanything} supports a broader set of geometric outputs. These methods obtain strong observed-surface priors from image- or voxel-aligned predictions, but their output density and coverage remain coupled to the input views. Our work studies whether a fixed set of explicit 3D queries can provide a complementary scene-level representation under sparse and uneven coverage.}

\vspace{5pt}
{\textbf{Anchor- and query-based 3D representations.}
Scaffold-GS~\cite{lu2024scaffold} uses optimized anchors to spawn local Gaussians for per-scene reconstruction. In multi-view 3D detection, DETR3D~\cite{wang2022detr3d} and PETR~\cite{liu2022petrpositionembeddingtransformation} associate explicit 3D object queries with image features, while Point-DETR3D~\cite{gao2024pointdetr3d} studies positional query initialization from point priors. These methods decode a sparse set of supervised objects rather than a dense renderable scene. LRM~\cite{hong2024lrm}, LGM~\cite{tang2024lgm}, and 3DShape2VecSet~\cite{zhang20233dshape2vecset} use learned tokens for object reconstruction or generation. LVSM~\cite{jin2024lvsm} and SceneTok~\cite{asim2026scenetok} more closely match the \emph{role} of a compact scene-token representation, but their tokens are latent and are decoded into target views or generative scene representations. In \ourmethod, each token carries an explicit 3D coordinate and spawns a local Gaussian cluster, while its DETR-like cross-attention supplies the \emph{mechanism} for collecting multi-view evidence.}

{Concurrent work explores complementary directions. AnchorSplat~\cite{zhang2026anchorsplat} uses 3D geometric anchors and a Gaussian refiner for scene-level feedforward reconstruction, whereas TokenGS~\cite{ren2026tokengs} decouples 3D Gaussian prediction from pixels with learnable implicit tokens. Tab.~\ref{tab:comparison} summarizes the factual representation differences most relevant to our setting.}

%% file: tables/comparison_of_representations.tex

\begin{table*}[!t]
    \centering
    \caption{
    Comparison of representations for 3D reconstruction. \cmark{} = favorable, \dash{} = moderate, \xmark{} = unfavorable. \emph{Volumetric}: can represent geometry beyond visible surfaces (not pixel-aligned). \emph{Fast Inference}: single forward pass without per-scene optimization. \emph{Low Memory}: compact representation. \ourmethod achieves the best overall trade-off.
    }
    \resizebox{1\linewidth}{!}{
    \begin{tabular}{cccccc}
    \toprule
    & Representation & Methods & Volumetric & Fast Inference & Low Memory \\
    \midrule
    Per-scene 
        & GS & 3D GS, Mip-Splatting & \cmark & \xmark & \dash \\
    Optimization    & NeRF & Mip-NeRF & \cmark  & \xmark & \cmark \\
    \midrule
    \multirow{4}{*}{Feedforward} 
        & Pixel-aligned Pointmaps & VGGT, Pi3 & \xmark & \cmark & \cmark \\
        & Voxel-aligned GS & AnySplat & \dash & \cmark & \dash \\
        & Voxel-aligned NeRF & DistillNeRF & \cmark & \xmark & \dash \\
        & Queries (Ours) & \textbf{\ourmethod} & \cmark & \cmark & \cmark \\
    \bottomrule
    \end{tabular}
    }
    \label{tab:comparison}
    \vspace{-16pt}
\end{table*}

%% file: sec/3_method.tex
\section{\ourmethod}

We propose \ourmethod, an efficient framework based on sparse queries for fast reconstruction from unposed images. We start with the problem statement and notations in Sec.~\ref{sec:notation}, followed by the network structure in Sec.~\ref{sec:network}, and the training procedure in Sec.~\ref{sec:training}.

\subsection{Notations}
\label{sec:notation}
{Given $N$ RGB images $\{\mathbf{I}_i\}_{i=1}^{N}$ of the same scene, with $\mathbf{I}_i \in \mathbb{R}^{3 \times H \times W}$, \ourmethod associates their features with $Q$ learnable queries $\mathcal{Q}=\{\mathbf{q}_i\}_{i=1}^{Q}$. Feedforward models predict camera-to-world transforms and per-view local point maps; applying each predicted transform to its local point map yields global points in one shared per-scene gauge. This coordinate system is non-metric and is not a first-camera frame. Queries, spawned Gaussians, and rendering cameras all use the same gauge, so a global similarity change does not alter the rendered supervision. Each query carries a location $\mathbf{p}_i$ in this frame and spawns $K$ Gaussians through learned local offsets. The resulting set $\mathcal{G}=\{\mathbf{g}_{ik}\mid i=1,\ldots,Q;\,k=1,\ldots,K\}$ contains $QK$ colored Gaussian primitives~\cite{kerbl2023gaussian} and is rendered as $\hat{\mathbf{I}}=\texttt{Render}(\mathcal{G},\pi)$ for a camera $\pi\in\mathbb{SE}(3)$.}

In addition, following the formulation of VGGT~\cite{wang2025vggt} and Pi3~\cite{wang2025pi3permutationequivariantvisualgeometry}, the image input is also mapped to a corresponding set of per-frame 3D annotations:
\begin{equation}
f\big(\{\mathbf{I}_i\}_{i=1}^{N}\big) = \{(\mathbf{P}_i, \mathbf{X}_i, \mathbf{C}_i)\}_{i=1}^{N},
\end{equation}
{where for each frame $\mathbf{I}_i$, the transformer predicts (1) a camera-to-world pose $\mathbf{P}_i \in \mathbb{SE}(3)$, (2) a local point map $\mathbf{X}_i \in \mathbb{R}^{3 \times H \times W}$, and (3) a confidence map $\mathbf{C}_i \in \mathbb{R}^{H \times W}$.}

\begin{figure*}[t]
    \centering
    \includegraphics[width=\linewidth]{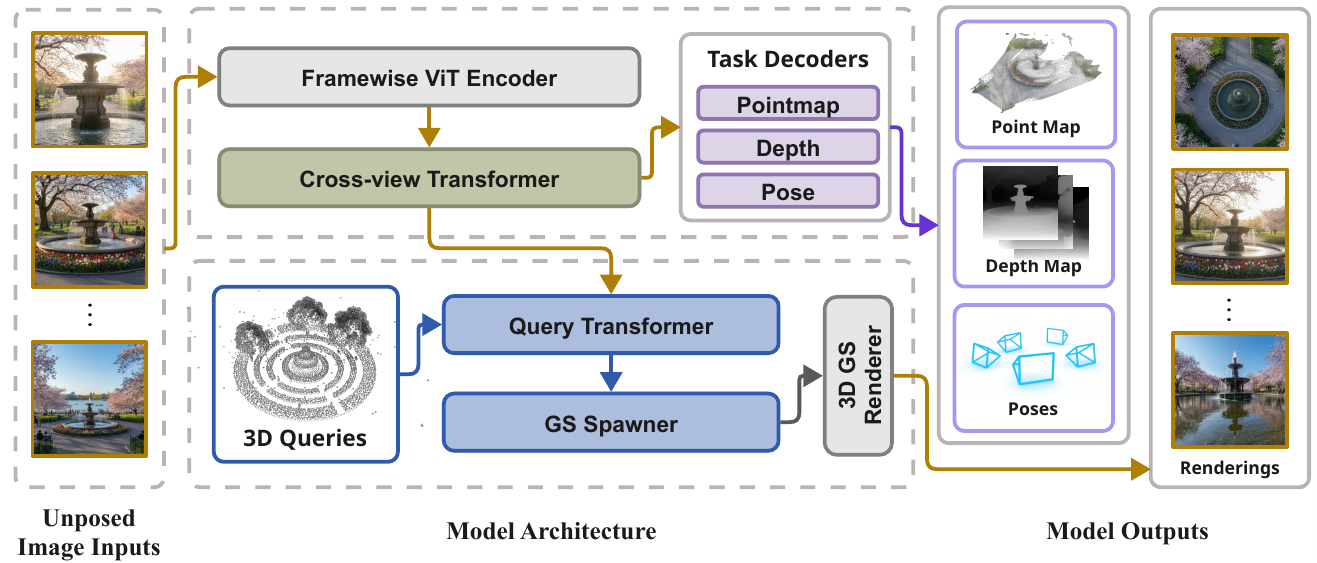}
    \caption{\textbf{\ourmethod pipeline overview.} Given multi-view images, a ViT encoder with alternating attention extracts per-frame tokens and decodes camera poses and point maps. A set of 3D queries is refined through cross-attention with image tokens and self-attention among queries. Each query then spawns $K$ Gaussians, which are rendered via differentiable splatting for RGB and depth supervision.}
    \label{fig:pipeline}
\end{figure*}

\subsection{Query-based Reconstructor}
\label{sec:network}
Rather than relying on dense 3D representations such as voxels or large sets of Gaussians, which are computationally expensive and scale poorly, we adopt a sparse set of learnable 3D queries as our core scene representation. Each query corresponds to a point in 3D space with an associated high-dimensional latent embedding. This sparse representation captures the coarse structure and appearance of the scene while enabling efficient feedforward inference. An overview of the \ourmethod pipeline is shown in Fig.~\ref{fig:pipeline}.

\vspace{5pt}
\noindent  \textbf{Image tokenization.} {Given $N$ unposed images $\mathbf{I}_i \in \mathbb{R}^{3 \times H \times W}$, we extract per-frame visual tokens with a Vision Transformer (ViT). We use DINOv2 as the backbone for its strong semantic representations:}
\begin{equation}
    \mathbf{T}_i = \texttt{DINO}(\mathbf{I}_i) \in {\mathbb{R}^{HW / p^2 \times d}}
\end{equation}
where $p$ is the patch size and $d$ is the token dimensionality.

To aggregate cross-view information, we follow VGGT~\cite{wang2025vggt} and Pi3~\cite{wang2025pi3permutationequivariantvisualgeometry} and apply an {alternating-attention (AA) transformer}, which alternates between intra-frame and inter-frame attention operations:
\begin{equation}
    \{\mathbf{T}_i^{l+1}\} = \texttt{AA-Transformer} (\{\mathbf{T}_i^{l}\})
\end{equation}
{where $l$ denotes the transformer-layer index.}

The aggregated image tokens serve two purposes. First, they are decoded into per-frame geometric annotations---camera poses $\mathbf{P}_i$, point maps $\mathbf{X}_i$, and confidence maps $\mathbf{C}_i$:
\begin{equation}
  \{\mathbf{P}_i, \mathbf{X}_i, \mathbf{C}_i\} = \texttt{Decoder} (\{\mathbf{T}_i\}) 
\end{equation}

These predicted geometric attributes provide strong {geometric priors} that guide the subsequent update of 3D queries, enabling more accurate and consistent reconstruction for the observed surfaces.

\vspace{5pt}
\noindent  \textbf{Query Initialization.} {Each query has an explicit 3D coordinate and latent embedding. Detection frameworks such as DETR~\cite{carion2020endtoendobjectdetectiontransformers} and DETR3D~\cite{wang2022detr3d} can learn sparse object queries with direct box supervision. In our dense reconstruction setting, random-only query coordinates are unstable because 2D rendered image losses rather than ground-truth 3D boxes are used during training.}

{To address this, we adopt a hybrid initialization strategy. Half of the query coordinates are sampled from the predicted non-metric point maps, providing geometric grounding on observed surfaces. The remaining learnable anchors are initialized uniformly in the normalized 3D scene range. Uniformity applies only to their initialization: after transformer refinement and the predicted query deformation, they are free to move away from the initial samples. The final deformed query is the center of a local Gaussian cluster, and neither the query centers nor the spawned Gaussians are constrained to be uniformly distributed on a surface.}

\begin{figure}
    \centering
\includegraphics[width=\linewidth]{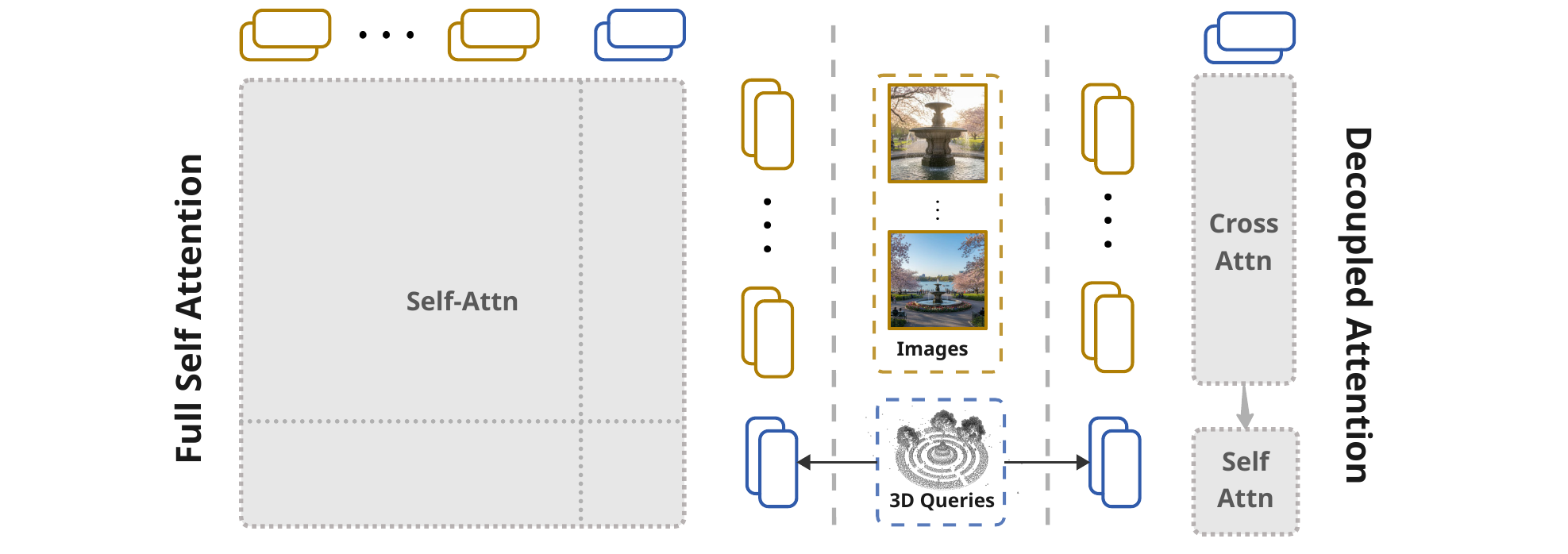}
    \caption{{Architectural comparison of full self-attention over concatenated image/query tokens and the adopted decoupled design with query-to-image cross-attention followed by query-only self-attention.}}
    \label{fig:attention}
\end{figure}

\noindent  \textbf{Query propagation.}
After query initialization, the image tokens from the tokenizer $\mathcal{T} = \{\mathbf{T}_i\}_{i=1}^N$, together with the query tokens $\mathcal{Q}=\{\mathbf{q}_i\}_{i=1}^Q$, are fed into the Query Transformer to update the query tokens:
\begin{equation}
    \mathcal{Q} = \textsc{QueryTransformer}(\mathcal{Q}, \mathcal{T})
\end{equation}
{We use a fixed budget of $Q=4096$ queries and $K=64$ Gaussians per query in all experiments except the query-count ablation in Fig.~\ref{fig:ablation1}. This budget is independent of the number and resolution of the input views, although the image-token count still grows linearly with the number of views.}

As illustrated in Fig.~\ref{fig:attention}, we consider two variants of the Query Transformer.
The most straightforward approach is to concatenate the query tokens and image tokens and apply full self-attention as follows: 
\begin{equation}
    \mathcal{Q}, \mathcal{T} = \textsc{Self-Attn}([\mathcal{Q}, \mathcal{T}])
\end{equation}
where the output image tokens will be discarded. 

However, this results in a computational complexity of $\mathcal{O}\left(\left(Q+NHW/p^2\right)^2\right)$, which is prohibitively expensive for high-resolution inputs. 

{We therefore use 12 Query Transformer layers with a decoupled attention design. In each layer, cross-attention first updates queries from image tokens, followed by self-attention among the queries:}
\begin{equation}
 \mathcal{Q} = \textsc{Cross-Attn}(\mathcal{Q}, \mathcal{T}), \mathcal{Q} = \textsc{Self-Attn}(\mathcal{Q})
\end{equation}
{Writing $T=HW/p^2$ for the number of tokens per image, this changes the theoretical attention complexity from $\mathcal{O}((Q+NT)^2)$ to $\mathcal{O}(QNT+Q^2)$, yielding substantial memory savings and enabling the model to scale up to larger size, high-resolution images and more queries. The end-to-end memory and timing comparison in Sec.~\ref{sec:experiments} also reflects the representation and decoder.}

For the image tokens, we convert the predicted camera poses into Plücker ray embeddings, which serve as positional encodings.
Each query naturally carries its own 3D coordinate as its positional embedding.
This 3D-aware positional parameterization enables more meaningful geometric interaction between image tokens and queries, improving spatial reasoning and reconstruction fidelity.

\vspace{5pt}
\noindent  \textbf{GS spawning.}
Without the reliance on ground truth 3D supervision, we leverage Gaussian Splatting as a differentiable bridge between 3D geometry and 2D observations. By rendering 3D primitives onto the image plane, we can supervise reconstruction using abundant 2D signals---RGB and depth maps---which are far more available than explicit 3D annotations.

{Each query $\mathbf{q}_i$ at location $\mathbf{p}_i$ spawns $K$ Gaussians $\{\mathbf{g}_{ik}\}_{k=1}^K$. Its updated embedding first predicts a deformation $\delta\mathbf{q}_i$, so $\mathbf{p}_i+\delta\mathbf{q}_i$ becomes the cluster center. An MLP then decodes $K$ local offsets $\delta\mathbf{g}_{ik}$ around that center, while attribute heads predict opacity, scale, rotation, and color.}
\begin{equation}
\mathcal{G} =
\bigcup_{i=1}^{Q}
\bigcup_{k=1}^{K}
\big(
\mathbf{p}_i + \delta \mathbf{q}_i + \delta \mathbf{g}_{ik},
\Sigma_{ik},
\mathbf{c}_{ik},
\alpha_{ik}
\big),
\end{equation}

The resulting Gaussian set $\mathcal{G}$ can then be rendered into RGB and depth efficiently via a GS rasterizer.
Our design preserves the efficiency of sparse 3D queries while enabling the spawned dense Gaussians to model fine-grained geometric and appearance details, yielding high-fidelity reconstructions at low computational cost.

\subsection{Training}
\label{sec:training}

\noindent \textbf{Losses.}
{The Gaussians $\mathcal{G}$ are rendered into RGB images $\hat{\mathbf{I}}$ and depth maps $\hat{\mathbf{D}}$ through differentiable splatting. In Stage 1, $\mathcal{L}_\text{rgb}$ includes an $\ell_1$ reconstruction term and an LPIPS perceptual term. For high-resolution Stage 2 fine-tuning, we omit the LPIPS term and use only the $\ell_1$ term because LPIPS substantially increases training time. Both stages also optimize the scale-invariant rendered-depth loss $\mathcal{L}_\text{depth}$ and the backbone camera loss $\mathcal{L}_\text{cam}$. The total loss is}
\begin{equation}
    \mathcal{L} = \mathcal{L}_\text{rgb} + \lambda_d \mathcal{L}_\text{depth} + \lambda_c \mathcal{L}_\text{cam}
    \label{eq:training_loss}
\end{equation}
{We set $\lambda_d=0.1$ and $\lambda_c=0.2$. Table~\ref{tab:ablation} evaluates the $224^2$ Stage 1 model, whereas Tables~\ref{tab:sparse_nvs} and~\ref{tab:dense_nvs} use the $448^2$ Stage 2 model.}

\vspace{5pt}
\noindent \textbf{Novel-view supervision.}
{The supervision views are a \emph{superset} of the input views. Given three inputs, for example, we supervise six renderings: the three inputs and three held-out training views. Missing appearance or geometry visible in those held-out targets contributes to the rendering loss, so this objective is intended to discourage reproducing only input-view surfaces. If the model only focuses on reconstructing the observed areas from the input views, empty holes will present in the novel viewpoints, and such artifacts will be corrected with the target novel images as supervision.}

\vspace{5pt}
\noindent \textbf{Staged training.}
{The DINOv2 encoder, alternating-attention transformer, and point-map head are initialized from Pi3~\cite{wang2025pi3permutationequivariantvisualgeometry} and kept frozen. We update the camera head, 12-layer Query Transformer, and Gaussian spawning/attribute heads, for 502.31M trainable parameters; the frozen backbone contains 890.99M parameters (1.393B total). We train with AdamW (learning rate $10^{-4}$ with cosine decay) for 100 epochs at $224^2$ on 32 A100 GPUs, then fine-tune for 20 epochs at $448^2$. Each sample contains 2--64 input views, while the query budget remains $Q=4096$.}

\subsection{Post Optimization}
\label{sec:optim}

Following AnySplat~\cite{jiang2025anysplat}, rendering quality can be further improved through test-time optimization (TTO). Using our predicted Gaussians as initialization and predicted camera poses as input, we render the Gaussians back into the input views and compute reconstruction losses against the original images. Backpropagating these losses refines the Gaussian parameters. We evaluate this protocol in Sec.~\ref{sec:experiments} under the dense-view setting.

%% file: sec/4_experiments.tex
\section{Experiments}
\label{sec:experiments}

\subsection{Experimental Setups}

\noindent \textbf{Datasets.} {We use the data-processing pipelines from MapAnything~\cite{keetha2025mapanything} and CUT3R~\cite{wang2025continuous3dperceptionmodel}. Training uses DL3DV-10K~\cite{ling2024dl3dv}, ScanNet++~\cite{yeshwanthliu2023scannetpp}, BlendedMVS~\cite{yao2020blendedmvs}, and Co3D~\cite{reizenstein21co3d}. We evaluate NVS on Mip-NeRF 360~\cite{barron2022mipnerf360unboundedantialiased}, VR-NeRF~\cite{Xu_2023}, and a 150-scene RealEstate10K~\cite{zhou2018stereo} subset. Camera-pose evaluation uses RealEstate10K and Co3Dv2~\cite{reizenstein21co3d}.}
\input{tables/sparse_nvs}

\input{tables/realestate_nvs}

\input{tables/pose_res_new}

\vspace{5pt}
\noindent \textbf{Evaluation Metrics and Protocol.} 
We adopt PSNR, SSIM \cite{1284395}, and LPIPS \cite{zhang2018perceptual} for novel-view synthesis, and depth abs-rel error for geometric accuracy. For pose estimation, we follow Pi3 \cite{wang2025pi3permutationequivariantvisualgeometry} using RRA, RTA, and AUC at a 30-degree threshold.

{For Mip-NeRF 360 and VR-NeRF sparse-view evaluation, input and test views are disjoint. We average five input-view subsets per scene and compute metrics only on their held-out targets. For RealEstate10K, we randomly sample 150 scenes from the test split used by NoPoSplat~\cite{ye2024poseproblemsurprisinglysimple}, with two input and two disjoint target views per scene. Every method uses the same scenes and view selections. Inference time is the end-to-end latency from images to Gaussians on one A100 80GB GPU.}

\vspace{5pt}
\noindent \textbf{Baselines.} We compare \ourmethod against a set of methods for 3D scene reconstruction, novel view synthesis, and camera pose estimation. For novel view synthesis, we include NoPoSplat~\cite{ye2024poseproblemsurprisinglysimple} and multi-view, transformer-based methods such as AnySplat~\cite{jiang2025anysplat}. To evaluate reconstruction quality, we also test the per-scene optimization performance using 3DGS~\cite{kerbl20233dgaussiansplattingrealtime} and MipSplatting~\cite{Yu2023MipSplatting} with different GS initializations and predicted camera poses from VGGT~\cite{wang2025vggt}, AnySplat~\cite{jiang2025anysplat}, and \ourmethod. For camera pose estimation, we benchmark against Fast3R~\cite{yang2025fast3r3dreconstruction1000}, CUT3R~\cite{wang2025continuous3dperceptionmodel}, FLARE~\cite{zhang2025flare}, VGGT~\cite{wang2025vggt}, and Pi3~\cite{wang2025pi3permutationequivariantvisualgeometry}.

\vspace{5pt}
\noindent \textbf{Implementation details.}
Training details are described in Sec.~\ref{sec:training}. We use $Q = 4096$ queries with $K = 64$ Gaussians per query by default, resulting in 262K total Gaussians. All runtime measurements use a single A100 80GB GPU with batch size 1. We follow MapAnything~\cite{keetha2025mapanything} for dynamic resolution during training with aspect ratios 1.0, 1.33, 1.52, 1.77, and 2.0.

\input{tables/dense_nvs}
\input{tables/efficiency}

\subsection{Quantitative Results}
{Following AnySplat~\cite{jiang2025anysplat}, we evaluate sparse inputs (3 or 6 views) and dense inputs (32 or 64 views). Tab.~\ref{tab:sparse_nvs} shows that \ourmethod obtains the best PSNR among the compared feedforward methods for both sparse-view benchmarks. The metric picture is mixed on VR-NeRF: AnySplat has higher SSIM and lower LPIPS for three views, and lower LPIPS for six views. Tab.~\ref{tab:realestate_nvs} summarizes the two-view RealEstate10K evaluation, where \ourmethod leads the compared methods on all three metrics.}

Under the dense-setting evaluation (Tab.~\ref{tab:dense_nvs}), our approach achieves competitive feedforward results and provides a significantly better initialization for per-scene optimization (3DGS+Ours, MipSplatting+Ours). We attribute the gap in the feedforward-only dense setting to the inherent sparsity of our representation: we use up to $4096 \times 64 = 262\text{K}$ Gaussians, significantly fewer than per-pixel methods which generate $448 \times 448 \times N$ primitives. Despite this, \ourmethod provides substantially better initialization quality, as shown by the large gains in the post-optimization rows.

The evaluation results on camera pose estimation are detailed in Tab.~\ref{tab:relpose-vggt}. The performance is comparable to Pi3~\cite{wang2025pi3permutationequivariantvisualgeometry}. We attribute the slight gap to more limited training data since Pi3 uses private dataset during the training.

It's noteworthy that in Tab.~\ref{tab:dense_nvs}, 3DGS+AnySplat occasionally underperforms standalone AnySplat on VR-NeRF. We attribute this to inaccurate pose estimates from AnySplat degrading the per-scene optimization, since the predicted Gaussians from AnySplat are aligned with the predicted camera poses. On the other hand, \ourmethod provides both more accurate poses and a better Gaussian initialization.

\vspace{5pt}
\noindent \textbf{Efficiency and geometry.}
{Tab.~\ref{tab:efficiency} compares end-to-end efficiency and rendered-depth accuracy. Relative to AnySplat, \ourmethod uses $14.7\times$ fewer Gaussians, 39\% less GPU memory, and $2.35\times$ less inference time, while reducing depth Abs Rel from 0.062 to 0.038.}

\subsection{Qualitative Analysis}

\begin{figure*}[t]
    \centering
    \includegraphics[width=\linewidth]{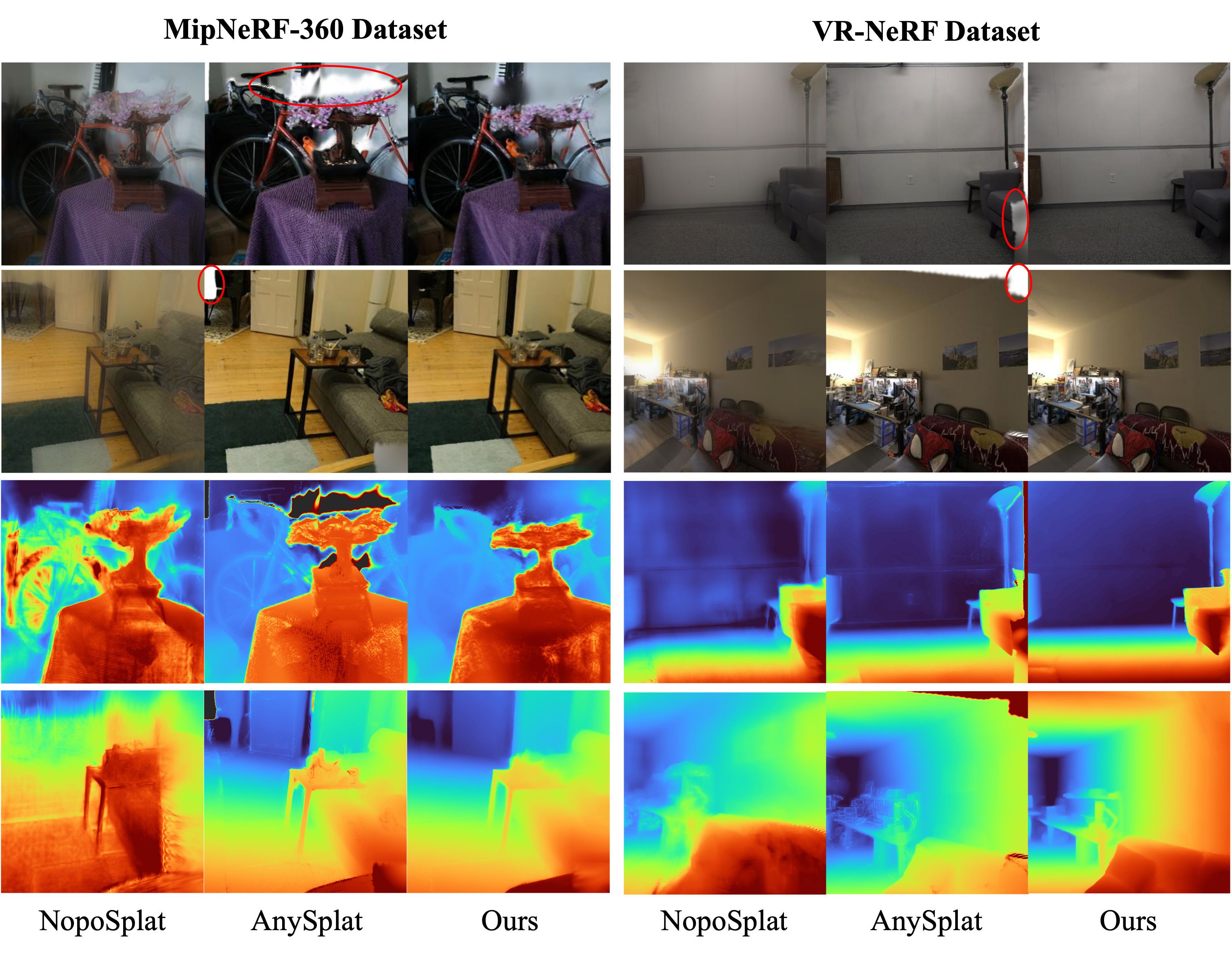} 
    \caption{{\textbf{Qualitative and geometric comparison.} Top: RGB rendered on held-out novel views. Bottom: depth rendered from each method's predicted Gaussians, not from our backbone point-map head. In the illustrated regions with limited input-view coverage, AnySplat contains blank RGB regions and depth holes, while \ourmethod produces fewer missing regions.}}
    \label{fig:qual}
\end{figure*}

Figure~\ref{fig:qual} presents qualitative comparisons of rendered novel views and depth maps. NoPoSplat struggles to produce high-fidelity outputs, often exhibiting severe artifacts and incomplete geometry. AnySplat, while more stable, frequently generates noticeable blank regions due to its reliance on per-pixel predictions, which cannot allocate Gaussians to unobserved areas. In contrast, \ourmethod delivers consistently higher-quality renderings with more complete structures and sharper details. The depth maps further confirm this: AnySplat exhibits holes and noisy depth in occluded regions, while \ourmethod produces cleaner and more complete geometry with sharper boundaries, consistent with the quantitative depth improvement in Tab.~\ref{tab:efficiency}.

\begin{figure*}[t]
    \centering
    \begin{subfigure}[b]{0.32\textwidth}
        \centering
        \includegraphics[width=\textwidth]{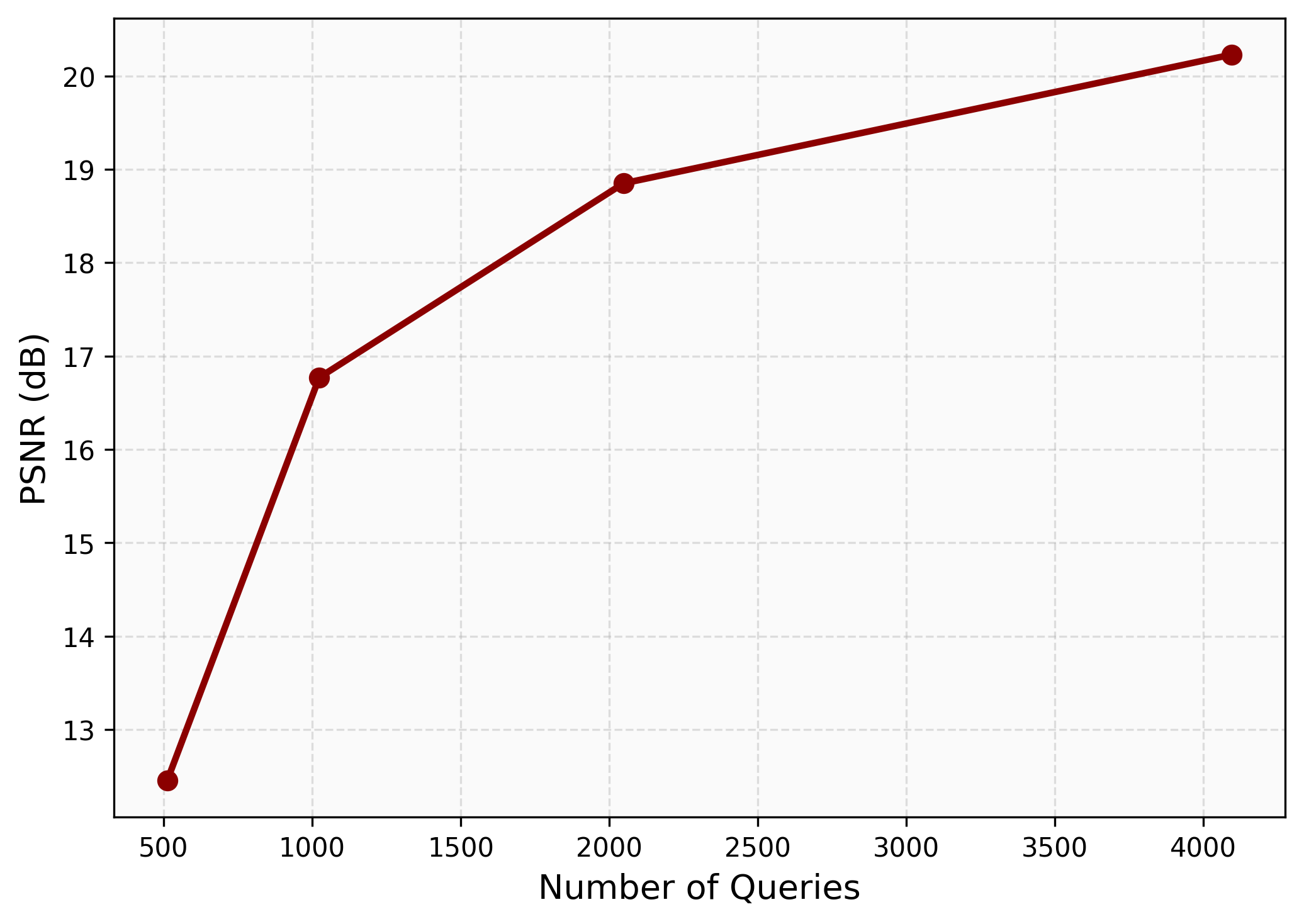}
        \caption{Ablation on number of Queries}
        \label{fig:ablation1}
    \end{subfigure}
    \hfill
    \begin{subfigure}[b]{0.32\textwidth}
        \centering
        \includegraphics[width=\textwidth]{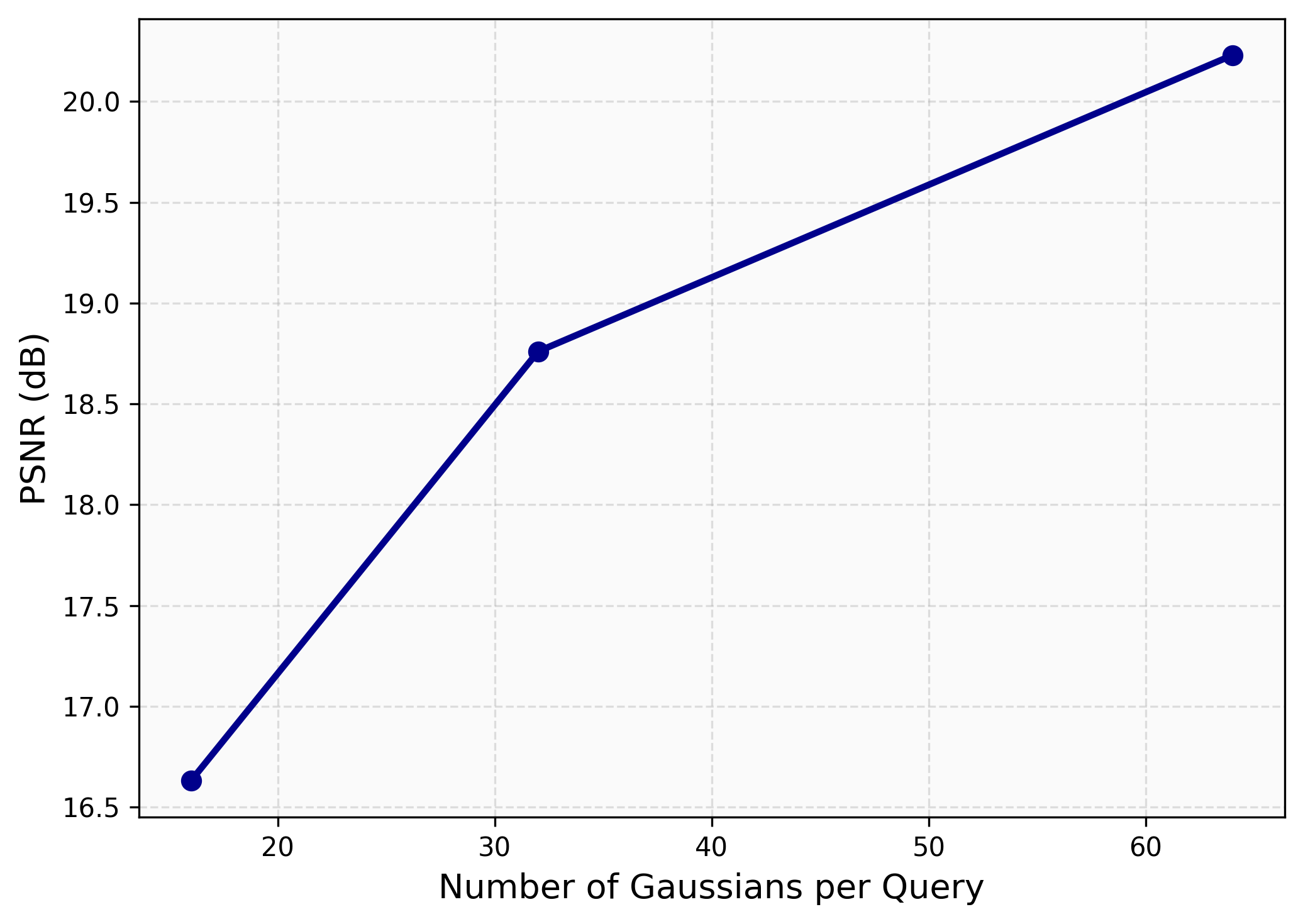}
        \caption{Ablation on number of Gaussians}
        \label{fig:ablation2}
    \end{subfigure}
    \hfill
    \begin{subfigure}[b]{0.32\textwidth}
        \centering
        \includegraphics[width=\textwidth]{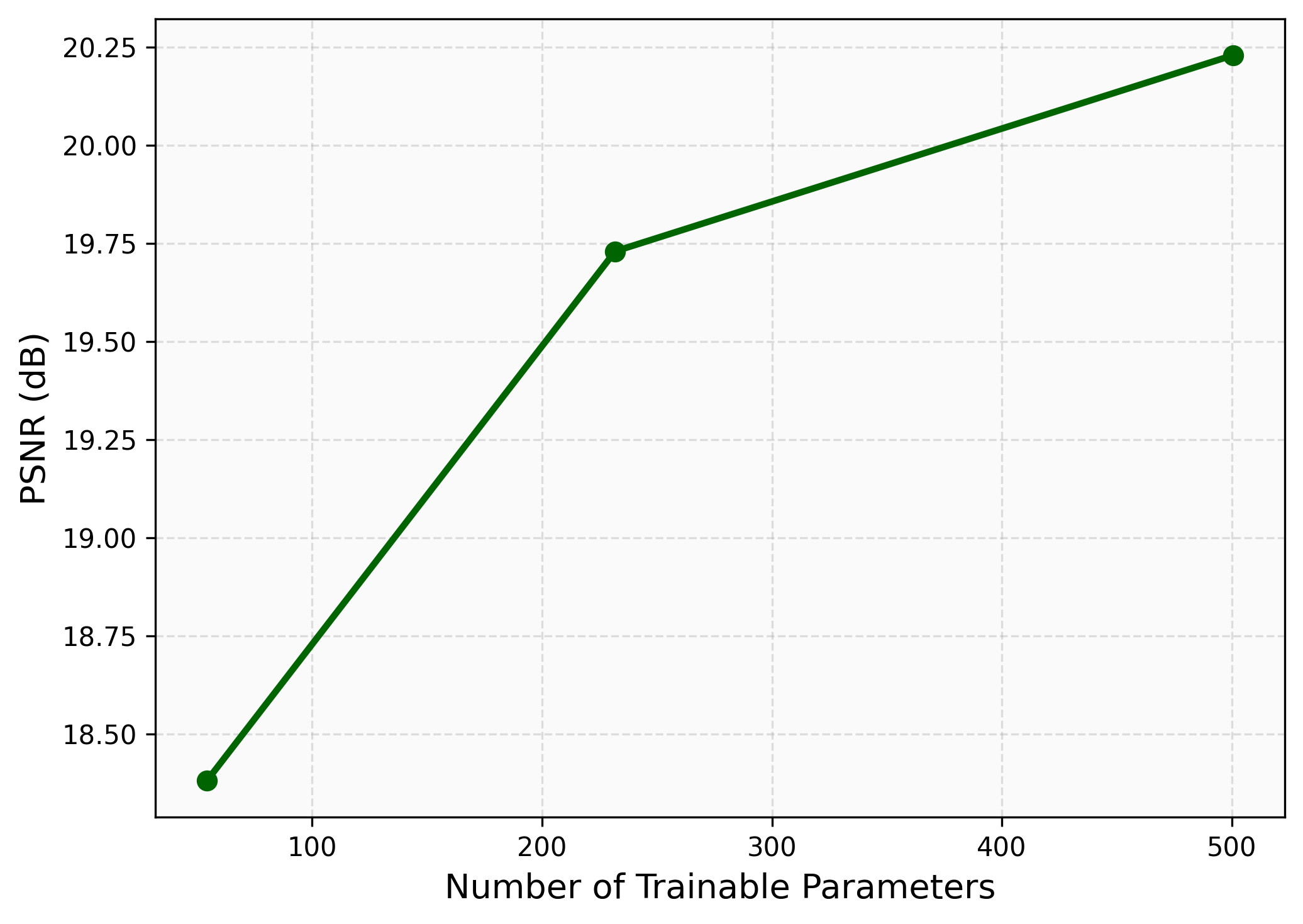}
        \caption{{Trainable parameters (M)}}
        \label{fig:ablation3}
    \end{subfigure}

    \caption{{Stage-1 ($224^2$) PSNR ablations on (a) query count, (b) Gaussians per query, and (c) trainable model capacity in millions. Increasing any of these factors leads to consistent improvements in PSNR, demonstrating clear scaling behavior in \ourmethod.}}
    \label{fig:three_in_a_row}
\end{figure*}

\begin{table} \small
    \centering
    \caption{{Stage-1 ($224^2$) ablations on Mip-NeRF 360 with three input views. }}
    \begin{tabular}{lccc}
    \toprule
        Configuration &PSNR $\uparrow$&  SSIM $\uparrow$&  LPIPS $\downarrow$\\
    \midrule
         (a) w/o depth rendering & 19.96 & 0.718 & 0.234 \\
        {(b) Point-map queries only} & {18.58} & {0.627} & {0.313} \\
         {(c) Random initialization only} & {12.11} & {0.259} & {0.574} \\
    \midrule
    {Full model (hybrid initialization)} & {20.23} & {0.713} & {0.182}\\
 \bottomrule
    \end{tabular}
    \label{tab:ablation}
\end{table}

\subsection{Ablation Studies}

We perform a series of ablations to assess the contribution of each component in \ourmethod as follows: 

\vspace{5pt}
\noindent \textbf{Number of queries.}
We begin by examining how the reconstruction quality scales with the number of queries. As shown in Fig.~\ref{fig:ablation1}, increasing the query count consistently boosts PSNR, indicating that a larger query set provides greater spatial coverage and representational capacity.

\vspace{5pt}
\noindent \textbf{Gaussians per query.}
Next, we vary the number of Gaussians (16, 32 and 64) spawned by each query with a total number of 4096 queries. Figure~\ref{fig:ablation2} shows that allocating more Gaussians per query results in higher PSNR, confirming that a richer local density representation enables finer detail synthesis.

\vspace{5pt}
\noindent \textbf{Model size.}
We further ablate the architecture size by adjusting the depth and hidden size of the transformer. As illustrated in Fig.~\ref{fig:ablation3}, models with larger capacity achieve noticeably better rendering quality, reflecting a clear scaling trend with respect to the trainable parameter count.

\vspace{5pt}
\noindent \textbf{Depth rendering.}
To understand the role of depth supervision, we disable Gaussian-rendered depth during training. As shown in Tab.~\ref{tab:ablation}(a), omitting depth rendering leads to a marked degradation in performance, highlighting the importance of depth cues for stabilizing geometry prediction.

\vspace{5pt}
\noindent \textbf{Query initialization.}
{Finally, we compare the two initialization sources in Tab.~\ref{tab:ablation}(b-c). Point-map-only initialization remains geometrically grounded but is weaker than the full hybrid model, while random-only initialization collapses. This confirms that grounding queries using coarse point-map estimates is crucial for stable optimization and high-quality rendering.}

%% file: tables/sparse_nvs.tex
\begin{table*}[t!] \small
\caption{{Sparse-view novel-view synthesis on held-out target views. We report PSNR, SSIM, LPIPS, and the inference time in seconds for a single feedforward pass from images to Gaussians. Best and second-best results are shown in \textbf{bold} and \underline{underline}, respectively.}}
\centering
\resizebox{1\linewidth}{!}{
\begin{tabular}{ll|cccc|cccc}
\toprule
\multirow{2}{*}{Dataset} & \multirow{2}{*}{Method} &
\multicolumn{4}{c|}{3 Views} &
\multicolumn{4}{c}{6 Views} \\
\cmidrule(lr){3-6}\cmidrule(lr){7-10}
& & PSNR$\uparrow$ & SSIM$\uparrow$ & LPIPS$\downarrow$ & Time (s)$\downarrow$
& PSNR$\uparrow$ & SSIM$\uparrow$ & LPIPS$\downarrow$ & Time (s)$\downarrow$ \\
\midrule

\multirow{3}{*}{Mip-NeRF}
& NoPoSplat~\cite{ye2024poseproblemsurprisinglysimple} & 18.21 & 0.482 & 0.426 & 0.416 & 16.06 & 0.423 & 0.540 & 2.121 \\
& AnySplat~\cite{jiang2025anysplat} & \underline{20.08} & \underline{0.606} & \underline{0.274} & \underline{0.279} & \underline{18.29} & \underline{0.518} & \underline{0.336} & \underline{0.646} \\
& \ourmethod & \textbf{22.70} & \textbf{0.660} & \textbf{0.261} & \textbf{0.213} & \textbf{21.80} & \textbf{0.622} & \textbf{0.300} & \textbf{0.447} \\

\midrule

\multirow{3}{*}{VR-NeRF}
& NoPoSplat~\cite{ye2024poseproblemsurprisinglysimple} & \underline{21.28} & \underline{0.744} & \underline{0.381} & 0.406 & \underline{20.69} & \textbf{0.728} & \underline{0.426} & 2.176 \\
& AnySplat~\cite{jiang2025anysplat} & 19.67 & \textbf{0.745} & \textbf{0.313} & \underline{0.290} & 17.25 & 0.683 & \textbf{0.411} & \underline{0.656} \\
& \ourmethod & \textbf{21.99} & 0.708 & 0.446 & \textbf{0.198} & \textbf{20.87} & \underline{0.689} & 0.431 & \textbf{0.412} \\

\bottomrule
\end{tabular}
}
\label{tab:sparse_nvs}
\end{table*}

%% file: tables/realestate_nvs.tex
\begin{table}[t]
    \centering
    \small
    \caption{{Novel-view synthesis on 150 randomly sampled scenes from the RealEstate10K test split used by NoPoSplat, with two input and two disjoint target views per scene. All methods use the same sampled scenes and views.}}
    \label{tab:realestate_nvs}
    {
    \begin{tabular}{lccc}
        \toprule
        Method & PSNR $\uparrow$ & SSIM $\uparrow$ & LPIPS $\downarrow$ \\
        \midrule
        FLARE~\cite{zhang2025flare} & 17.74 & 0.586 & 0.371 \\
        AnySplat~\cite{jiang2025anysplat} & 18.93 & 0.615 & 0.226 \\
        \ourmethod & \textbf{20.50} & \textbf{0.672} & \textbf{0.196} \\
        \bottomrule
    \end{tabular}}
\end{table}

%% file: tables/pose_res_new.tex
\begin{table*}[t] \small
    \centering
    \caption{\textbf{Camera Pose Estimation on RealEstate10K~\cite{zhou2018stereo} and Co3Dv2~\cite{reizenstein21co3d}.}
    }
    \resizebox{1.0\linewidth}{!}{
    \begin{tabular}{lcccccc}
        \toprule
        \multicolumn{1}{l}{\multirow{3}{*}{\textbf{Method}}} &
        \multicolumn{3}{c}{\textbf{RealEstate10K}} &
        \multicolumn{3}{c}{\textbf{Co3Dv2}} \\
        \cmidrule(r){2-4} \cmidrule(r){5-7}
        \multicolumn{1}{c}{} &
        RRA@30 $\uparrow$ & RTA@30 $\uparrow$ & AUC@30 $\uparrow$ &
        RRA@30 $\uparrow$ & RTA@30 $\uparrow$ & AUC@30 $\uparrow$ \\
        \midrule
        Fast3R~\cite{yang2025fast3r3dreconstruction1000} & 99.05 & 81.86 & 61.68 & 97.49 & 91.11 & 73.43\\
        CUT3R~\cite{wang2025continuous3dperceptionmodel} & 99.82 & 95.10 & {81.47} & 96.19 & 92.69 & 75.82 \\
        FLARE~\cite{zhang2025flare} & 99.69 & {95.23} & 80.01 & 96.38 & 93.76 & 73.99 \\
        VGGT~\cite{wang2025vggt} & \underline{99.97} & 93.13 & 77.62 & {98.96} & {97.13} & \textbf{88.59} \\
        {Pi3 }~\cite{wang2025pi3permutationequivariantvisualgeometry} & \textbf{99.99} & \textbf{95.62} & \textbf{85.90} & \textbf{99.05} & \underline{97.33} & {88.41} \\
        \ourmethod & \textbf{99.99} & \underline{95.44} & \underline{83.69} & \textbf{99.05} & \textbf{97.44} & \underline{88.52} \\
        \bottomrule
    \end{tabular}
    }
    \label{tab:relpose-vggt}
\end{table*}

%% file: tables/dense_nvs.tex
\begin{table*}[t!] \small
\caption{Quantitative comparison for dense-view novel-view synthesis. We report PSNR, SSIM, and LPIPS. Best and second-best are in \textbf{bold} and \underline{underline}. Top rows: feedforward only. Bottom rows: feedforward initialization followed by per-scene 3DGS/MipSplatting optimization.}
\centering
\resizebox{1\linewidth}{!}{
\begin{tabular}{ll|cccccc}
\toprule
 \multirow{2}{*}{Dataset} &\multirow{2}{*}{Method} &
\multicolumn{3}{c}{32 Views}&
\multicolumn{3}{c}{64 Views}\\
& & PSNR$\uparrow$ & SSIM$\uparrow$ & LPIPS$\downarrow$ & PSNR$\uparrow$ & SSIM$\uparrow$ & LPIPS$\downarrow$ \\
\midrule

\multirow{8}{*}{{Mip-NeRF 360}} &AnySplat~\cite{jiang2025anysplat} & \textbf{22.32} & \textbf{0.660} & \textbf{0.258} & 21.26 & 0.607 & \textbf{0.303} \\

 &Ours & 21.51 & 0.619 & 0.325 & \textbf{21.58} & \textbf{0.641} & 0.335 \\

\cmidrule(lr){2-8}

 &3DGS+VGGT~\cite{kerbl20233dgaussiansplattingrealtime, wang2025vggt} & 22.61 & 0.701 & 0.233 & \underline{23.78} & \underline{0.729} & \underline{0.221} \\

 &3DGS+AnySplat~\cite{kerbl20233dgaussiansplattingrealtime, jiang2025anysplat} & \underline{24.99} & \underline{0.705} & \underline{0.227} & 23.71 & 0.664 & 0.266 \\

 &3DGS+Ours~\cite{kerbl20233dgaussiansplattingrealtime} & \textbf{25.14} & \textbf{0.774} & \textbf{0.183} & \textbf{26.00} & \textbf{0.784} & \textbf{0.176} \\

\cmidrule(lr){2-8}
 &MipSplatting+VGGT~\cite{Yu2023MipSplatting, wang2025vggt} & 22.27&0.686&0.247&23.72&\underline{0.721}&\underline{0.230} \\

 &MipSplatting+AnySplat~\cite{Yu2023MipSplatting, jiang2025anysplat} & \underline{24.98}&\underline{0.710}&\underline{0.223}&\underline{23.84}&0.675&0.257 \\

 &MipSplatting+Ours~\cite{Yu2023MipSplatting} & \textbf{25.26}&\textbf{0.772}&\textbf{0.184}&\textbf{25.99}&\textbf{0.782}&\textbf{0.178} \\

\midrule

\multirow{8}{*}{VR-NeRF} &AnySplat~\cite{jiang2025anysplat} & \textbf{23.21}&\textbf{0.773}&\textbf{0.308}&23.62&0.802&\textbf{0.285} \\

 &Ours & 23.18&0.770&0.395&\textbf{24.74}&\textbf{0.808}&0.334 \\

\cmidrule(lr){2-8}

 &3DGS+VGGT~\cite{kerbl20233dgaussiansplattingrealtime, wang2025vggt} & 21.74&\underline{0.720}&\underline{0.352}&23.18&0.748&0.315 \\

 &3DGS+AnySplat~\cite{kerbl20233dgaussiansplattingrealtime, jiang2025anysplat} & \underline{21.90}&0.708&0.377&\underline{24.27}&\underline{0.768}&\underline{0.306} \\

 &3DGS+Ours~\cite{kerbl20233dgaussiansplattingrealtime} & \textbf{27.03}&\textbf{0.846}&\textbf{0.223}&\textbf{28.56}&\textbf{0.879}&\textbf{0.174} \\

\cmidrule(lr){2-8}

 &MipSplatting+VGGT~\cite{Yu2023MipSplatting, wang2025vggt} & 21.79&\underline{0.715}&\underline{0.362}&23.50&0.759&\underline{0.308} \\

 &MipSplatting+AnySplat~\cite{Yu2023MipSplatting, jiang2025anysplat} & \underline{21.93}&0.711&0.373&\underline{24.08}&\underline{0.764}&0.311 \\

 &MipSplatting+Ours~\cite{Yu2023MipSplatting} & \textbf{26.62}&\textbf{0.838}&\textbf{0.237}&\textbf{28.39}&\textbf{0.873}&\textbf{0.185} \\

\bottomrule
\end{tabular}
}
\label{tab:dense_nvs}
\end{table*}

%% file: tables/efficiency.tex
\begin{table}[t] \small
\caption{{End-to-end efficiency and rendered-depth comparison with 32 input views at $448^2$ on one A100 80GB (batch size 1). UniQueR uses $14.7\times$ fewer Gaussians, 39\% less memory, and $2.35\times$ less time than AnySplat.}}
\centering
\begin{tabular}{lcccc}
\toprule
Method & \# Gaussians & GPU Mem. & Time (s) & Depth Abs Rel $\downarrow$ \\
\midrule
AnySplat & 3.85M & 18.42 GB & 4.63 & 0.062 \\
\ourmethod & $\leq$ {262K} & 11.19 GB & 1.97 & 0.038 \\
\bottomrule
\end{tabular}
\label{tab:efficiency}
\end{table}

%% file: sec/5_conclusion.tex
\section{Conclusion}

We presented \ourmethod, a unified query-based feedforward framework for 3D reconstruction from unposed images. Unlike existing feedforward methods that rely on depth maps or pixel-aligned Gaussians, \ourmethod introduces learnable 3D queries that decouple the scene representation from input viewpoints. Each query spawns a set of Gaussians supervised through differentiable rendering of RGB and depth, requiring no ground-truth 3D data. Experiments on Mip-NeRF 360 and VR-NeRF demonstrate improvements in both rendering quality and geometric accuracy over prior feedforward approaches, while using significantly fewer primitives and less memory.

\noindent \textbf{Limitations.}
{the current formulation assumes static scenes and does not model temporal dynamics. The fixed query budget can underrepresent highly complex scenes and trades some high-frequency perceptual detail for compactness. The model also depends on a large frozen pretrained geometry backbone. Adaptive query allocation, lighter backbones, controlled component ablations, and temporally persistent queries are promising directions.}

%% file: sec/X_supplement.tex
\clearpage
\appendix

\section{Implementation Details}
\label{sec:suppl_impl}

This section provides additional details beyond those in Sec.~3 of the main paper. Unless otherwise stated, all results reported for \ourmethod correspond to the single feedforward model described in the main paper. Rows labeled 3DGS+Ours or MipSplatting+Ours denote standard per-scene optimization initialized from our predicted Gaussian centers and camera poses.

\vspace{5pt}
\textbf{Training Datasets and Curation}
Following the data-processing pipeline from MapAnything~\cite{keetha2025mapanything}, we use its implementation for DL3DV-10K, ScanNet++ and BlendedMVS datasets. We follow CUT3R~\cite{wang2025continuous3dperceptionmodel} for the processing of Co3D. 
The evaluation benchmarks of MipNeRF-360 and VR-NeRF are curated by AnySplat~\cite{jiang2025anysplat}. For VR-NeRF sparse-view evaluation, AnySplat only provides curated subsets for the dense-view setting; we therefore curate subsets of 3 and 6 input views following the same protocol as Mip-NeRF 360, selecting 5 different versions and reporting averaged results. For the few-view settings, the rendering performance varies significantly across view selections, which motivates averaging over multiple subsets.

\vspace{5pt}
\textbf{Model Configuration}
Unless otherwise stated, we use up to $Q = 4096$ queries and spawn $K = 64$ Gaussians per query, leading to up to 262K Gaussians in total. The cross-view transformer and point-map head are initialized from Pi3~\cite{wang2025pi3permutationequivariantvisualgeometry} and kept frozen. We train the camera head and the query transformer, which preserves strong surface-level geometric priors while allowing the model to allocate additional Gaussians to regions not well explained by pixel-aligned predictions.
Half the queries are initialized from predicted pointmaps (non-metric). The other half are learnable spatial anchors initialized uniformly in $[-1.2, 1.2]^3$, scaled according to the predicted point range during training.

\vspace{5pt}
\textbf{Training Schedule and Supervision}
We follow MapAnything~\cite{keetha2025mapanything} for dynamic resolution during training. We first train on resolution $224$ with aspect ratios 1.0, 1.33, 1.52, 1.77, and 2.0, then fine-tune at resolution $448$. Training uses AdamW with learning rate $1 \times 10^{-4}$ and cosine decay for 100 epochs at $224$ and 20 additional epochs at $448$ on $32 \times$ A100 GPUs. During training, we randomly sample 2--64 input views per scene and the number of queries varies accordingly. Supervision views are a superset of the input views: besides reconstructing the input images, the predicted Gaussians are also rendered into held-out views and supervised there. This explicitly penalizes empty holes in unseen regions and encourages the queries to model geometry beyond the observed surfaces.

\vspace{5pt}
\textbf{Losses}
The spawned Gaussians are rendered into RGB images and depth maps with a differentiable Gaussian rasterizer. We supervise the model with an RGB reconstruction loss composed of $\ell_1$ and LPIPS, a scale-invariant depth loss on the rendered depth, and the camera loss inherited from the backbone. The depth term is important in practice because it regularizes the spatial layout of the spawned Gaussians and reduces floaters and holes in difficult views.

\subsection{Hyperparameter Summary}
Tab.~\ref{tab:suppl_hparams} summarizes the main hyperparameters used in the current version. 
\begin{table}[t]
    \centering
    \caption{Key hyperparameters used by \ourmethod.}
    \label{tab:suppl_hparams}
    \small
    \begin{tabular}{lc}
        \toprule
        Item & Value \\
        \midrule
        Number of queries $Q$ & 4096 \\
        Gaussians per query $K$ & 64 \\
        Total Gaussians & $\leq 262$K \\
        Optimizer & AdamW \\
        Initial learning rate & $1 \times 10^{-4}$ \\
        Training schedule & 100 epochs @ 224, 20 epochs @ 448 \\
        Input views per scene & 2--64 \\
        Dynamic aspect ratios & 1.0, 1.33, 1.52, 1.77, 2.0 \\
        Training hardware & $32\times$ A100 GPUs \\
        Timing hardware & single A100 80GB GPU \\
        \bottomrule
    \end{tabular}
\end{table}

\subsection{Evaluation Protocol}
\label{sec:suppl_eval}

\noindent\textbf{Sparse-view novel-view synthesis.}
For sparse-view evaluation, the input views and the testing views are strictly disjoint. We measure PSNR, SSIM, and LPIPS only on held-out views. To reduce variance from view sampling, we evaluate 5 curated subsets per scene and report the average. For the 3-view setting, a typical subset contains 5 frames: 3 are used as inputs and the remaining 2 are held out for evaluation.
For the 6-view setting, a typical subset contains 12 frames: 6 are used as inputs and the remaining 6 are held out for evaluation.

\vspace{5pt}
\noindent\textbf{Dense-view setting.}
For the 32-view and 64-view experiments, we follow AnySplat~\cite{jiang2025anysplat} and evaluate both the direct feedforward predictions and per-scene optimization initialized from each method. In the latter case, the optimizer receives the predicted camera poses together with the predicted Gaussian or point-based initialization. This protocol isolates the quality of the initialization and explains why stronger camera estimates lead to better downstream optimization.

\vspace{5pt}
\noindent\textbf{Geometry metric.}
To measure geometric accuracy, we render the predicted Gaussians into depth maps under the evaluation cameras and compute the absolute relative error (Abs Rel) against ground-truth depth. The numbers reported in Tab.~\ref{tab:efficiency} of the main paper use 32 input views at resolution $448 \times 448$. Because the depth is produced by the same Gaussian representation used for rendering RGB, this metric directly reflects whether the method reconstructs complete and geometrically consistent 3D structure.


\subsection{Post Optimization Protocol}
For post-optimization evaluation (Tab.~2 of the main paper), we use the predicted poses from VGGT, AnySplat, and \ourmethod. Since VGGT predicts points rather than full Gaussians, we use its points directly, while for AnySplat and \ourmethod we retain only the Gaussian centers and colors for initialization. The downstream optimization stage itself is identical across methods; only the initialization and camera poses differ.

\section{Additional Results}
\label{sec:suppl_results}

\subsection{Occlusion Handling and Geometry Completion}
\begin{figure}[t]
    \centering
    \includegraphics[width=\linewidth]{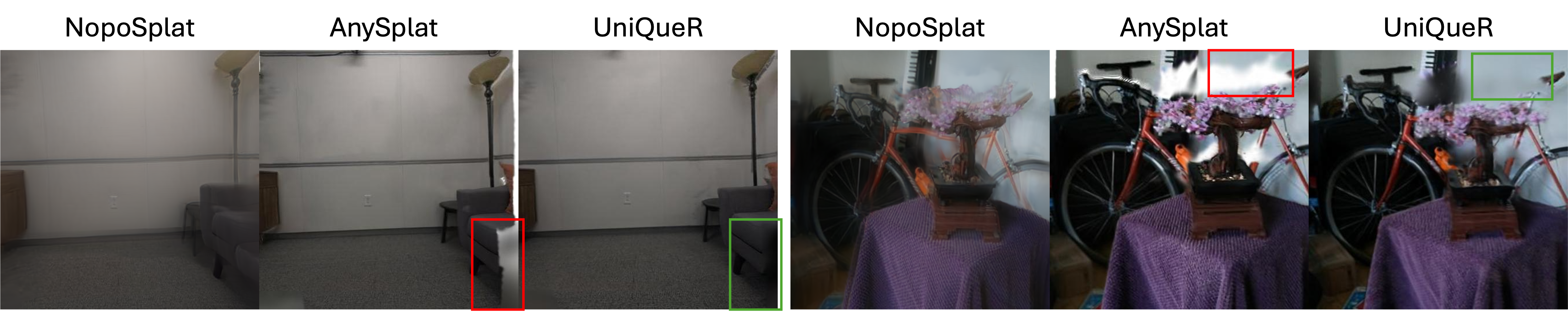}
    \caption{\textbf{Occlusion handling comparison.} Pixel-aligned predictions tend to leave empty holes in unseen regions, while \ourmethod can place Gaussians in occluded space and render more complete novel views.}
    \label{fig:occ_supp}
\end{figure}

A central claim of \ourmethod is that the query-based representation can place Gaussians outside the visible input surfaces. Fig.~\ref{fig:occ_supp} shows a representative example. Methods based on pixel-aligned predictions often leave empty holes when a region is not supported by the input image grids. In contrast, our sparse 3D queries are free to move in global 3D space and are trained with held-out-view supervision, which encourages them to allocate structure in occluded regions. This behavior is also reflected by the lower depth error reported in the main paper.

\subsection{Qualitative Visualizations}

\begin{figure}[t]
    \centering
    \includegraphics[width=0.7\linewidth]{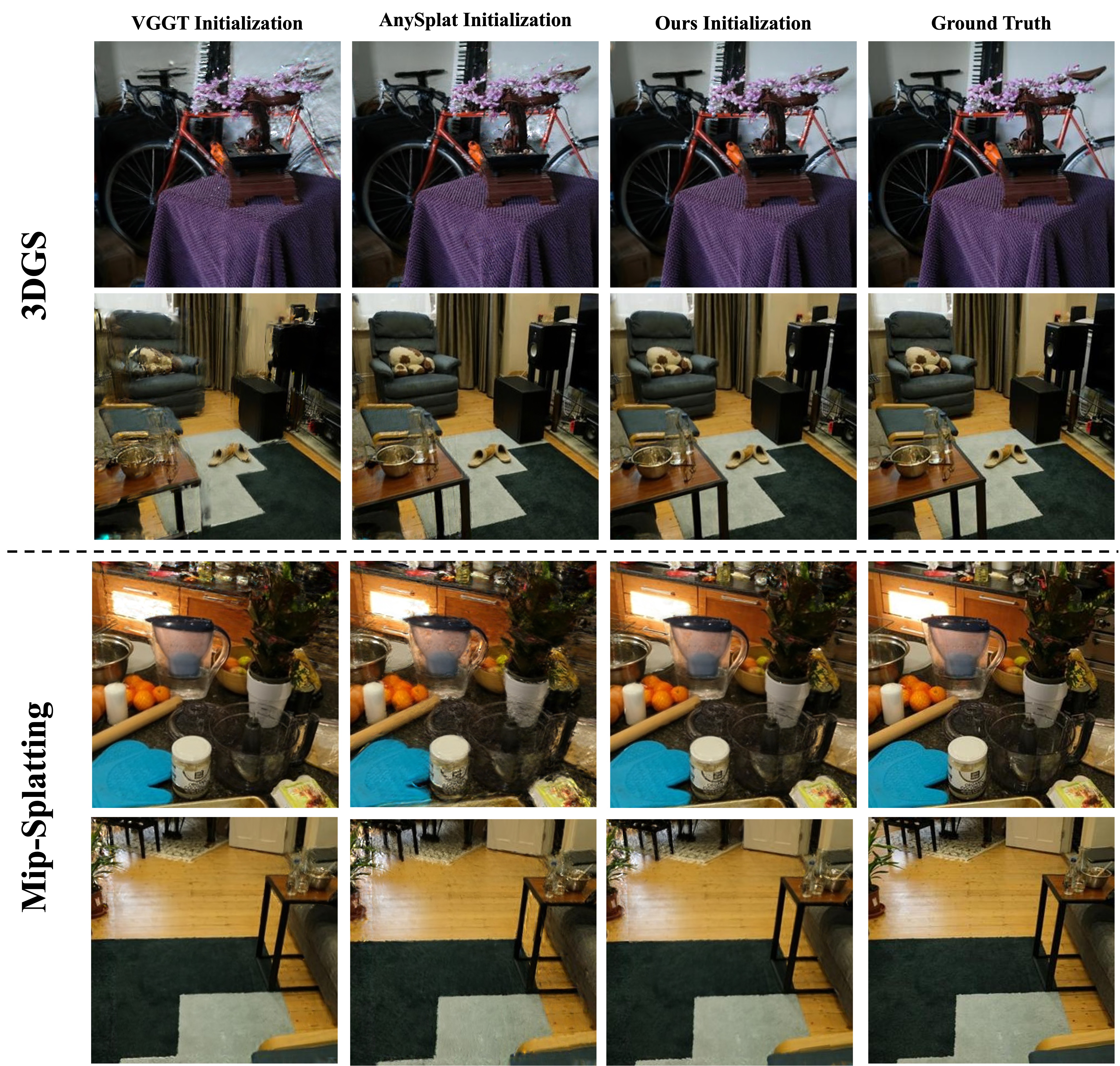}
    \caption{3DGS and Mip-Splatting post-optimization with initializations from various methods.}
    \label{fig:qual_3dgs_mip}
\end{figure}

Fig.~\ref{fig:qual_3dgs_mip} compares different Gaussian-splatting initialization strategies under both the 3DGS and Mip-Splatting rendering pipelines. Using VGGT or AnySplat as initialization often leads to blurred geometry, color inconsistencies, and structural drift. In contrast, our initialization produces sharper reconstructions, more stable geometry, and improved color fidelity across diverse scenes.



\section{Training Pipeline}
\label{sec:suppl_algo}

\begin{algorithm}[H]
\caption{\textbf{UniQueR Training Pipeline}}
\begin{algorithmic}[1]
    \Require Multi-view images $\{\mathbf{I}_i\}$
    \State $\mathbf{T}_i \gets \text{DINO}(\mathbf{I}_i)$
    \State $\{\mathbf{T}_i\} \gets \textsc{AA-Transformer}(\{\mathbf{T}_i\})$
    \State Predict $\{\mathbf{P}_i, \mathbf{X}_i, \mathbf{C}_i\} = \textsc{Decoder} (\{\mathbf{T}_i\}) $
    \State $Q \gets \textsc{InitQueries}(\mathbf{P}_i)$
    \For{$\ell=1..L$}
        \State $Q \gets \textsc{Cross-Attn}(Q, \{\mathbf{T}_i\})$
        \State $Q \gets \textsc{Self-Attn}(Q)$
    \EndFor
    \State $\mathcal{G} \gets \textsc{SpawnGaussians}(Q)$
    \State Render: $\hat{I}, \hat{D} \gets \textsc{Render}(\mathcal{G})$
    \State Compute losses and update parameters
\end{algorithmic}
\end{algorithm}

%% file: main.bib
@String(CVPR= {IEEE Conf. Comput. Vis. Pattern Recog.})

@String(ICCV= {Int. Conf. Comput. Vis.})

@String(ECCV= {Eur. Conf. Comput. Vis.})

@String(NIPS= {Adv. Neural Inform. Process. Syst.})

@String(TOG= {ACM Trans. Graph.})

@String(AAAI = {AAAI})

@String(CVPR  = {CVPR})

@String(ICCV  = {ICCV})

@String(ECCV  = {ECCV})

@String(NIPS  = {NeurIPS})

@String(TOG   = {ACM TOG})

@inproceedings{ren2026tokengs,
  title={Tokengs: Decoupling 3d gaussian prediction from pixels with learnable tokens},
  author={Ren, Jiawei and Tyszkiewicz, Michal Jan and Huang, Jiahui and Gojcic, Zan},
  booktitle={Proceedings of the IEEE/CVF Conference on Computer Vision and Pattern Recognition},
  pages={15365--15375},
  year={2026}
}

@inproceedings{ling2024dl3dv,
  title={Dl3dv-10k: A large-scale scene dataset for deep learning-based 3d vision},
  author={Ling, Lu and Sheng, Yichen and Tu, Zhi and Zhao, Wentian and Xin, Cheng and Wan, Kun and Yu, Lantao and Guo, Qianyu and Yu, Zixun and Lu, Yawen and others},
  booktitle={Proceedings of the IEEE/CVF Conference on Computer Vision and Pattern Recognition},
  pages={22160--22169},
  year={2024}
}

@misc{yang2025fast3r3dreconstruction1000,
      title={Fast3R: Towards 3D Reconstruction of 1000+ Images in One Forward Pass}, 
      author={Jianing Yang and Alexander Sax and Kevin J. Liang and Mikael Henaff and Hao Tang and Ang Cao and Joyce Chai and Franziska Meier and Matt Feiszli},
      year={2025},
      eprint={2501.13928},
      archivePrefix={arXiv},
      primaryClass={cs.CV},
      url={https://arxiv.org/abs/2501.13928}, 
}

@misc{wang2025continuous3dperceptionmodel,
      title={Continuous 3D Perception Model with Persistent State}, 
      author={Qianqian Wang and Yifei Zhang and Aleksander Holynski and Alexei A. Efros and Angjoo Kanazawa},
      year={2025},
      eprint={2501.12387},
      archivePrefix={arXiv},
      primaryClass={cs.CV},
      url={https://arxiv.org/abs/2501.12387}, 
}

@inproceedings{yang2024carff,
  title={CARFF: Conditional Auto-encoded Radiance Field for 3D Scene Forecasting},
  author={Yang, Jiezhi and Desai, Khushi and Packer, Charles and Bhatia, Harshil and Rhinehart, Nicholas and McAllister, Rowan and Gonzalez, Joseph E},
  booktitle={European Conference on Computer Vision},
  pages={225--242},
  year={2024},
  organization={Springer}
}

@ARTICLE{1284395,
  author={Zhou Wang and Bovik, A.C. and Sheikh, H.R. and Simoncelli, E.P.},
  journal={IEEE Transactions on Image Processing}, 
  title={Image quality assessment: from error visibility to structural similarity}, 
  year={2004},
  volume={13},
  number={4},
  pages={600-612},
  doi={10.1109/TIP.2003.819861}}

@misc{wang2025pi3permutationequivariantvisualgeometry,
      title={$\pi^3$: Permutation-Equivariant Visual Geometry Learning}, 
      author={Yifan Wang and Jianjun Zhou and Haoyi Zhu and Wenzheng Chang and Yang Zhou and Zizun Li and Junyi Chen and Jiangmiao Pang and Chunhua Shen and Tong He},
      year={2025},
      eprint={2507.13347},
      archivePrefix={arXiv},
      primaryClass={cs.CV},
      url={https://arxiv.org/abs/2507.13347}, 
}

@article{snavely2006photo,
  title={Photo tourism: exploring photo collections in 3D},
  author={Snavely, Noah and Seitz, Steven M and Szeliski, Richard},
  journal={TOG},
  volume={25},
  number={3},
  pages={835--846},
  year={2006},
}

@article{davison2007monoslam,
  title={MonoSLAM: Real-time single camera SLAM},
  author={Davison, Andrew J and Reid, Ian D and Molton, Nicholas D and Stasse, Olivier},
  journal={TPAMI},
  volume={29},
  number={6},
  pages={1052--1067},
  year={2007},
  publisher={IEEE}
}

@inproceedings{klein2007parallel,
  title={Parallel Tracking and Mapping for Small AR Workspaces},
  author={Klein, Georg and Murray, David},
  booktitle={ISMAR},
  pages={1--10},
  year={2007}
}

@inproceedings{agarwal2009building,
  title={Building Rome in a day},
  author={Agarwal, Sameer and Snavely, Noah and Simon, Ian and Seitz, Steven M and Szeliski, Richard},
  booktitle={ICCV},
  pages={72--79},
  year={2009},
}

@article{furukawa2009furu,
  title={Accurate, dense, and robust multiview stereopsis},
  author={Furukawa, Yasutaka and Ponce, Jean},
  journal={TPAMI},
  volume={32},
  number={8},
  pages={1362--1376},
  year={2009},
}

@inproceedings{triggs2000bundle,
  title={Bundle adjustment---a modern synthesis},
  author={Triggs, Bill and McLauchlan, Philip F and Hartley, Richard I and Fitzgibbon, Andrew W},
  booktitle={ICCVW},
  pages={298--372},
  year={2000},
}

@inproceedings{agarwal2010bundle,
  title={Bundle adjustment in the large},
  author={Agarwal, Sameer and Snavely, Noah and Seitz, Steven M and Szeliski, Richard},
  booktitle={ECCV},
  pages={29--42},
  year={2010},
}

@inproceedings{wu2011multicore,
  title={Multicore bundle adjustment},
  author={Wu, Changchang and Agarwal, Sameer and Curless, Brian and Seitz, Steven M},
  booktitle={CVPR},
  pages={3057--3064},
  year={2011},
}

@inproceedings{crandall2011discrete,
  title={Discrete-continuous optimization for large-scale structure from motion},
  author={Crandall, David and Owens, Andrew and Snavely, Noah and Huttenlocher, Dan},
  booktitle={CVPR},
  pages={3001--3008},
  year={2011},
}

@inproceedings{newcombe2011dtam,
  title={DTAM: Dense tracking and mapping in real-time},
  author={Newcombe, Richard A and Lovegrove, Steven J and Davison, Andrew J},
  booktitle={ICCV},
  pages={2320--2327},
  year={2011},
}

@inproceedings{wu2013towards,
  title={Towards linear-time incremental structure from motion},
  author={Wu, Changchang},
  booktitle={3DV},
  pages={127--134},
  year={2013},
}

@inproceedings{wilson2014robust,
  title={Robust global translations with 1dsfm},
  author={Wilson, Kyle and Snavely, Noah},
  booktitle={ECCV},
  pages={61--75},
  year={2014},
}

@inproceedings{sweeney2015optimizing,
  title={Optimizing the viewing graph for structure-from-motion},
  author={Sweeney, Chris and Sattler, Torsten and Hollerer, Tobias and Turk, Matthew and Pollefeys, Marc},
  booktitle={ICCV},
  pages={801--809},
  year={2015}
}

@inproceedings{galliani2015massively,
  title={Massively parallel multiview stereopsis by surface normal diffusion},
  author={Galliani, Silvano and Lasinger, Katrin and Schindler, Konrad},
  booktitle={ICCV},
  pages={873--881},
  year={2015}
}

@inproceedings{schonberger2016pixelwise,
  title={Pixelwise view selection for unstructured multi-view stereo},
  author={Sch{\"o}nberger, Johannes L and Zheng, Enliang and Frahm, Jan-Michael and Pollefeys, Marc},
  booktitle={ECCV},
  pages={501--518},
  year={2016},
}

@inproceedings{zhang2018perceptual,
  title={The Unreasonable Effectiveness of Deep Features as a Perceptual Metric},
  author={Zhang, Richard and Isola, Phillip and Efros, Alexei A and Shechtman, Eli and Wang, Oliver},
  booktitle={CVPR},
  year={2018}
}

@inproceedings{schonberger2016colmap,
  title={Structure-from-motion revisited},
  author={Schonberger, Johannes L and Frahm, Jan-Michael},
  booktitle={CVPR},
  pages={4104--4113},
  year={2016}
}

@inproceedings{bloesch2018codeslam,
  title={Codeslam---learning a compact, optimisable representation for dense visual slam},
  author={Bloesch, Michael and Czarnowski, Jan and Clark, Ronald and Leutenegger, Stefan and Davison, Andrew J},
  booktitle={CVPR},
  pages={2560--2568},
  year={2018}
}

@inproceedings{yao2018mvsnet,
  title={Mvsnet: Depth inference for unstructured multi-view stereo},
  author={Yao, Yao and Luo, Zixin and Li, Shiwei and Fang, Tian and Quan, Long},
  booktitle={ECCV},
  pages={767--783},
  year={2018}
}

@inproceedings{yao2020blendedmvs,
  title={Blendedmvs: A large-scale dataset for generalized multi-view stereo networks},
  author={Yao, Yao and Luo, Zixin and Li, Shiwei and Zhang, Jingyang and Ren, Yufan and Zhou, Lei and Fang, Tian and Quan, Long},
  booktitle={CVPR},
  pages={1790--1799},
  year={2020}
}

@inproceedings{teed2020raft,
  title={Raft: Recurrent all-pairs field transforms for optical flow},
  author={Teed, Zachary and Deng, Jia},
  booktitle={ECCV},
  pages={402--419},
  year={2020},
}

@inproceedings{mildenhall2020nerf,
  title={NeRF: Representing Scenes as Neural Radiance Fields for View Synthesis},
  author={Mildenhall, Ben and Srinivasan, Pratul P and Tancik, Matthew and Barron, Jonathan T and Ramamoorthi, Ravi and Ng, Ren},
  booktitle={ECCV},
  pages={405--421},
  year={2020},
}

@inproceedings{duzceker2021deepvideomvs,
  title={Deepvideomvs: Multi-view stereo on video with recurrent spatio-temporal fusion},
  author={Duzceker, Arda and Galliani, Silvano and Vogel, Christoph and Speciale, Pablo and Dusmanu, Mihai and Pollefeys, Marc},
  booktitle={CVPR},
  pages={15324--15333},
  year={2021}
}

@inproceedings{jang2021codenerf,
  title={Codenerf: Disentangled neural radiance fields for object categories},
  author={Jang, Wonbong and Agapito, Lourdes},
  booktitle={ICCV},
  pages={12949--12958},
  year={2021}
}

@inproceedings{wang2021neus,
  title={NeuS: learning neural implicit surfaces by volume rendering for multi-view reconstruction},
  author={Wang, Peng and Liu, Lingjie and Liu, Yuan and Theobalt, Christian and Komura, Taku and Wang, Wenping},
  booktitle={NIPS},
  pages={27171--27183},
  year={2021}
}

@inproceedings{yariv2021volsdf,
  title={Volume rendering of neural implicit surfaces},
  author={Yariv, Lior and Gu, Jiatao and Kasten, Yoni and Lipman, Yaron},
  booktitle={NIPS},
  pages={4805--4815},
  year={2021}
}

@inproceedings{barron2022mip,
  title={Mip-nerf 360: Unbounded anti-aliased neural radiance fields},
  author={Barron, Jonathan T and Mildenhall, Ben and Verbin, Dor and Srinivasan, Pratul P and Hedman, Peter},
  booktitle={CVPR},
  pages={5470--5479},
  year={2022}
}

@inproceedings{sayed2022simplerecon,
  title={Simplerecon: 3d reconstruction without 3d convolutions},
  author={Sayed, Mohamed and Gibson, John and Watson, Jamie and Prisacariu, Victor and Firman, Michael and Godard, Cl{\'e}ment},
  booktitle={ECCV},
  pages={1--19},
  year={2022},
}

@inproceedings{dexheimer2023learning,
  title={Learning a depth covariance function},
  author={Dexheimer, Eric and Davison, Andrew J},
  booktitle={CVPR},
  pages={13122--13131},
  year={2023}
}

@article{kerbl2023gaussian,
  title={3D Gaussian Splatting for Real-Time Radiance Field Rendering.},
  author={Kerbl, Bernhard and Kopanas, Georgios and Leimk{\"u}hler, Thomas and Drettakis, George},
  journal={TOG},
  volume={42},
  number={4},
  pages={139--1},
  year={2023}
}

@inproceedings{barron2023zip,
  title={Zip-nerf: Anti-aliased grid-based neural radiance fields},
  author={Barron, Jonathan T and Mildenhall, Ben and Verbin, Dor and Srinivasan, Pratul P and Hedman, Peter},
  booktitle={ICCV},
  pages={19697--19705},
  year={2023}
}

@inproceedings{doersch2023tapir,
  title={Tapir: Tracking any point with per-frame initialization and temporal refinement},
  author={Doersch, Carl and Yang, Yi and Vecerik, Mel and Gokay, Dilara and Gupta, Ankush and Aytar, Yusuf and Carreira, Joao and Zisserman, Andrew},
  booktitle={ICCV},
  pages={10061--10072},
  year={2023}
}

@inproceedings{yeshwanthliu2023scannetpp,
  title={ScanNet++: A High-Fidelity Dataset of 3D Indoor Scenes},
  author={Yeshwanth, Chandan and Liu, Yueh-Cheng and Nie{\ss}ner, Matthias and Dai, Angela},
  booktitle = {ICCV},
  year={2023}
}

@inproceedings{yin2023metric3d,
  title={Metric3d: Towards zero-shot metric 3d prediction from a single image},
  author={Yin, Wei and Zhang, Chi and Chen, Hao and Cai, Zhipeng and Yu, Gang and Wang, Kaixuan and Chen, Xiaozhi and Shen, Chunhua},
  booktitle={ICCV},
  pages={9043--9053},
  year={2023}
}

@article{karaev2023cotracker,
  title={Cotracker: It is better to track together},
  author={Karaev, Nikita and Rocco, Ignacio and Graham, Benjamin and Neverova, Natalia and Vedaldi, Andrea and Rupprecht, Christian},
  journal={arXiv preprint arXiv:2307.07635},
  year={2023}
}

@inproceedings{ke2024repurposing,
  title={Repurposing diffusion-based image generators for monocular depth estimation},
  author={Ke, Bingxin and Obukhov, Anton and Huang, Shengyu and Metzger, Nando and Daudt, Rodrigo Caye and Schindler, Konrad},
  booktitle={CVPR},
  pages={9492--9502},
  year={2024}
}

@inproceedings{xiao2024spatialtracker,
  title={SpatialTracker: Tracking Any 2D Pixels in 3D Space},
  author={Xiao, Yuxi and Wang, Qianqian and Zhang, Shangzhan and Xue, Nan and Peng, Sida and Shen, Yujun and Zhou, Xiaowei},
  booktitle={CVPR},
  pages={20406--20417},
  year={2024}
}

@inproceedings{huang20242d,
  title={2d gaussian splatting for geometrically accurate radiance fields},
  author={Huang, Binbin and Yu, Zehao and Chen, Anpei and Geiger, Andreas and Gao, Shenghua},
  booktitle={ACM SIGGRAPH},
  pages={1--11},
  year={2024}
}

@inproceedings{jang2024nvist,
  title={NViST: In the Wild New View Synthesis from a Single Image with Transformers},
  author={Jang, Wonbong and Agapito, Lourdes},
  booktitle={CVPR},
  pages={10181--10193},
  year={2024}
}

@misc{kerbl20233dgaussiansplattingrealtime,
      title={3D Gaussian Splatting for Real-Time Radiance Field Rendering}, 
      author={Bernhard Kerbl and Georgios Kopanas and Thomas Leimkühler and George Drettakis},
      year={2023},
      eprint={2308.04079},
      archivePrefix={arXiv},
      primaryClass={cs.GR},
      url={https://arxiv.org/abs/2308.04079}, 
}

@misc{wang20243dreconstructionspatialmemory,
      title={3D Reconstruction with Spatial Memory}, 
      author={Hengyi Wang and Lourdes Agapito},
      year={2024},
      eprint={2408.16061},
      archivePrefix={arXiv},
      primaryClass={cs.CV},
      url={https://arxiv.org/abs/2408.16061}, 
}

@misc{ye2024poseproblemsurprisinglysimple,
      title={No Pose, No Problem: Surprisingly Simple 3D Gaussian Splats from Sparse Unposed Images}, 
      author={Botao Ye and Sifei Liu and Haofei Xu and Xueting Li and Marc Pollefeys and Ming-Hsuan Yang and Songyou Peng},
      year={2024},
      eprint={2410.24207},
      archivePrefix={arXiv},
      primaryClass={cs.CV},
      url={https://arxiv.org/abs/2410.24207}, 
}

@misc{smart2024splatt3rzeroshotgaussiansplatting,
      title={Splatt3R: Zero-shot Gaussian Splatting from Uncalibrated Image Pairs}, 
      author={Brandon Smart and Chuanxia Zheng and Iro Laina and Victor Adrian Prisacariu},
      year={2024},
      eprint={2408.13912},
      archivePrefix={arXiv},
      primaryClass={cs.CV},
      url={https://arxiv.org/abs/2408.13912}, 
}

@misc{pan2024globalstructurefrommotionrevisited,
      title={Global Structure-from-Motion Revisited}, 
      author={Linfei Pan and Dániel Baráth and Marc Pollefeys and Johannes L. Schönberger},
      year={2024},
      eprint={2407.20219},
      archivePrefix={arXiv},
      primaryClass={cs.CV},
      url={https://arxiv.org/abs/2407.20219}, 
}

@misc{leroy2024groundingimagematching3d,
      title={Grounding Image Matching in 3D with MASt3R}, 
      author={Vincent Leroy and Yohann Cabon and Jérôme Revaud},
      year={2024},
      eprint={2406.09756},
      archivePrefix={arXiv},
      primaryClass={cs.CV},
      url={https://arxiv.org/abs/2406.09756}, 
}

@misc{charatan2024pixelsplat3dgaussiansplats,
      title={pixelSplat: 3D Gaussian Splats from Image Pairs for Scalable Generalizable 3D Reconstruction}, 
      author={David Charatan and Sizhe Li and Andrea Tagliasacchi and Vincent Sitzmann},
      year={2024},
      eprint={2312.12337},
      archivePrefix={arXiv},
      primaryClass={cs.CV},
      url={https://arxiv.org/abs/2312.12337}, 
}

@inproceedings{wang2024dust3r,
  title={Dust3r: Geometric 3d vision made easy},
  author={Wang, Shuzhe and Leroy, Vincent and Cabon, Yohann and Chidlovskii, Boris and Revaud, Jerome},
  booktitle={CVPR},
  pages={20697--20709},
  year={2024}
}

@article{Yu2023MipSplatting,
  author    = {Yu, Zehao and Chen, Anpei and Huang, Binbin and Sattler, Torsten and Geiger, Andreas},
  title     = {Mip-Splatting: Alias-free 3D Gaussian Splatting},
  journal   = {Conference on Computer Vision and Pattern Recognition (CVPR)},
  year      = {2024},
}

@article{chen2024mvsplat,
    title   = {MVSplat: Efficient 3D Gaussian Splatting from Sparse Multi-View Images},
    author  = {Chen, Yuedong and Xu, Haofei and Zheng, Chuanxia and Zhuang, Bohan and Pollefeys, Marc and Geiger, Andreas and Cham, Tat-Jen and Cai, Jianfei},
    journal = {arXiv preprint arXiv:2403.14627},
    year    = {2024},
}

@misc{szymanowicz2024splatterimageultrafastsingleview,
      title={Splatter Image: Ultra-Fast Single-View 3D Reconstruction}, 
      author={Stanislaw Szymanowicz and Christian Rupprecht and Andrea Vedaldi},
      year={2024},
      eprint={2312.13150},
      archivePrefix={arXiv},
      primaryClass={cs.CV},
      url={https://arxiv.org/abs/2312.13150}, 
}

@article{yu2024sim,
  title={SIM-Sync: From certifiably optimal synchronization over the 3D similarity group to scene reconstruction with learned depth},
  author={Yu, Xihang and Yang, Heng},
  journal={IEEE Robotics and Automation Letters},
  year={2024},
  publisher={IEEE}
}

@article{zhou2018stereo,
  title={Stereo magnification: Learning view synthesis using multiplane images},
  author={Zhou, Tinghui and Tucker, Richard and Flynn, John and Fyffe, Graham and Snavely, Noah},
  journal={arXiv preprint arXiv:1805.09817},
  year={2018}
}

@article{cabon2025must3r,
  title={MUSt3R: Multi-view Network for Stereo 3D Reconstruction},
  author={Cabon, Yohann and Stoffl, Lucas and Antsfeld, Leonid and Csurka, Gabriela and Chidlovskii, Boris and Revaud, Jerome and Leroy, Vincent},
  journal={arXiv preprint arXiv:2503.01661},
  year={2025}
}

@InProceedings{10.1007/978-3-031-73024-5_14,
author="Yang, Jiezhi
and Desai, Khushi
and Packer, Charles
and Bhatia, Harshil
and Rhinehart, Nicholas
and McAllister, Rowan
and Gonzalez, Joseph E.",
editor="Leonardis, Ale{\v{s}}
and Ricci, Elisa
and Roth, Stefan
and Russakovsky, Olga
and Sattler, Torsten
and Varol, G{\"u}l",
title="CARFF: Conditional Auto-Encoded Radiance Field for 3D Scene Forecasting",
booktitle="Computer Vision -- ECCV 2024",
year="2025",
publisher="Springer Nature Switzerland",
address="Cham",
pages="225--242",
isbn="978-3-031-73024-5"
}

@article{mueller2022instant,
    author = {Thomas M\"uller and Alex Evans and Christoph Schied and Alexander Keller},
    title = {Instant Neural Graphics Primitives with a Multiresolution Hash Encoding},
    journal = {ACM Trans. Graph.},
    issue_date = {July 2022},
    volume = {41},
    number = {4},
    month = jul,
    year = {2022},
    pages = {102:1--102:15},
    articleno = {102},
    numpages = {15},
    url = {https://doi.org/10.1145/3528223.3530127},
    doi = {10.1145/3528223.3530127},
    publisher = {ACM},
    address = {New York, NY, USA},
}

@misc{chen2024pref3rposefreefeedforward3d,
      title={PreF3R: Pose-Free Feed-Forward 3D Gaussian Splatting from Variable-length Image Sequence}, 
      author={Zequn Chen and Jiezhi Yang and Heng Yang},
      year={2024},
      eprint={2411.16877},
      archivePrefix={arXiv},
      primaryClass={cs.CV},
      url={https://arxiv.org/abs/2411.16877}, 
}

@inproceedings{wang2025vggt,
  title={VGGT: Visual Geometry Grounded Transformer},
  author={Wang, Jianyuan and Chen, Minghao and Karaev, Nikita and Vedaldi, Andrea and Rupprecht, Christian and Novotny, David},
  booktitle={Proceedings of the IEEE/CVF Conference on Computer Vision and Pattern Recognition},
  year={2025}
}

@article{jiang2025anysplat,
  title={AnySplat: Feed-forward 3D Gaussian Splatting from Unconstrained Views},
  author={Jiang, Lihan and Mao, Yucheng and Xu, Linning and Lu, Tao and Ren, Kerui and Jin, Yichen and Xu, Xudong and Yu, Mulin and Pang, Jiangmiao and Zhao, Feng and others},
  journal={arXiv preprint arXiv:2505.23716},
  year={2025}
}

@misc{keetha2025mapanything,
  title={{MapAnything}: Universal Feed-Forward Metric {3D} Reconstruction},
  author={Nikhil Keetha and Norman M\"{u}ller and Johannes Sch\"{o}nberger and Lorenzo Porzi and Yuchen Zhang and Tobias Fischer and Arno Knapitsch and Duncan Zauss and Ethan Weber and Nelson Antunes and Jonathon Luiten and Manuel Lopez-Antequera and Samuel Rota Bul\`{o} and Christian Richardt and Deva Ramanan and Sebastian Scherer and Peter Kontschieder},
  note={arXiv preprint arXiv:2509.13414},
  year={2025}
}

@misc{carion2020endtoendobjectdetectiontransformers,
      title={End-to-End Object Detection with Transformers}, 
      author={Nicolas Carion and Francisco Massa and Gabriel Synnaeve and Nicolas Usunier and Alexander Kirillov and Sergey Zagoruyko},
      year={2020},
      eprint={2005.12872},
      archivePrefix={arXiv},
      primaryClass={cs.CV},
      url={https://arxiv.org/abs/2005.12872}, 
}

@misc{liu2022petrpositionembeddingtransformation,
      title={PETR: Position Embedding Transformation for Multi-View 3D Object Detection}, 
      author={Yingfei Liu and Tiancai Wang and Xiangyu Zhang and Jian Sun},
      year={2022},
      eprint={2203.05625},
      archivePrefix={arXiv},
      primaryClass={cs.CV},
      url={https://arxiv.org/abs/2203.05625}, 
}

@misc{wang2023exploringobjectcentrictemporalmodeling,
      title={Exploring Object-Centric Temporal Modeling for Efficient Multi-View 3D Object Detection}, 
      author={Shihao Wang and Yingfei Liu and Tiancai Wang and Ying Li and Xiangyu Zhang},
      year={2023},
      eprint={2303.11926},
      archivePrefix={arXiv},
      primaryClass={cs.CV},
      url={https://arxiv.org/abs/2303.11926}, 
}

@inproceedings{zhang2025flare,
  title={Flare: Feed-forward geometry, appearance and camera estimation from uncalibrated sparse views},
  author={Zhang, Shangzhan and Wang, Jianyuan and Xu, Yinghao and Xue, Nan and Rupprecht, Christian and Zhou, Xiaowei and Shen, Yujun and Wetzstein, Gordon},
  booktitle={Proceedings of the Computer Vision and Pattern Recognition Conference},
  pages={21936--21947},
  year={2025}
}

@inproceedings{herau2023moisst,
  title={Moisst: Multimodal optimization of implicit scene for spatiotemporal calibration},
  author={Herau, Quentin and Piasco, Nathan and Bennehar, Moussab and Roldao, Luis and Tsishkou, Dzmitry and Migniot, Cyrille and Vasseur, Pascal and Demonceaux, C{\'e}dric},
  booktitle={IEEE/RSJ International Conference on Intelligent Robots and Systems (IROS)},
  pages={1810--1817},
  year={2023},
}

@inproceedings{herau2024soac,
  title={Soac: Spatio-temporal overlap-aware multi-sensor calibration using neural radiance fields},
  author={Herau, Quentin and Piasco, Nathan and Bennehar, Moussab and Roldao, Luis and Tsishkou, Dzmitry and Migniot, Cyrille and Vasseur, Pascal and Demonceaux, C{\'e}dric},
  booktitle={CVPR},
  pages={15131--15140},
  year={2024}
}

@inproceedings{herau20243dgs,
  title={3dgs-calib: 3d gaussian splatting for multimodal spatiotemporal calibration},
  author={Herau, Quentin and Bennehar, Moussab and Moreau, Arthur and Piasco, Nathan and Rold{\~a}o, Luis and Tsishkou, Dzmitry and Migniot, Cyrille and Vasseur, Pascal and Demonceaux, C{\'e}dric},
  booktitle={IEEE/RSJ International Conference on Intelligent Robots and Systems (IROS)},
  pages={8315--8321},
  year={2024},
}

@article{herau2025pose,
  title={Pose Optimization for Autonomous Driving Datasets using Neural Rendering Models},
  author={Herau, Quentin and Piasco, Nathan and Bennehar, Moussab and Rold{\~a}o, Luis and Tsishkou, Dzmitry and Liu, Bingbing and Migniot, Cyrille and Vasseur, Pascal and Demonceaux, C{\'e}dric},
  journal={arXiv preprint arXiv:2504.15776},
  year={2025}
}

@misc{barron2022mipnerf360unboundedantialiased,
      title={Mip-NeRF 360: Unbounded Anti-Aliased Neural Radiance Fields}, 
      author={Jonathan T. Barron and Ben Mildenhall and Dor Verbin and Pratul P. Srinivasan and Peter Hedman},
      year={2022},
      eprint={2111.12077},
      archivePrefix={arXiv},
      primaryClass={cs.CV},
      url={https://arxiv.org/abs/2111.12077}, 
}

@inproceedings{Xu_2023, series={SA '23},
   title={VR-NeRF: High-Fidelity Virtualized Walkable Spaces},
   url={http://dx.doi.org/10.1145/3610548.3618139},
   DOI={10.1145/3610548.3618139},
   booktitle={SIGGRAPH Asia 2023 Conference Papers},
   publisher={ACM},
   author={Xu, Linning and Agrawal, Vasu and Laney, William and Garcia, Tony and Bansal, Aayush and Kim, Changil and Rota Bul{\`o}, Samuel and Porzi, Lorenzo and Kontschieder, Peter and Bo{\v{z}}i{\v{c}}, Alja{\v{z}} and Lin, Dahua and Zollh{\"o}fer, Michael and Richardt, Christian},
   year={2023},
   month=dec, pages={1--12},
   collection={SA '23} }

@inproceedings{reizenstein21co3d,
	Author = {Reizenstein, Jeremy and Shapovalov, Roman and Henzler, Philipp and Sbordone, Luca and Labatut, Patrick and Novotny, David},
	Booktitle = {International Conference on Computer Vision},
	Title = {Common Objects in 3D: Large-Scale Learning and Evaluation of Real-life 3D Category Reconstruction},
	Year = {2021},
}

@inproceedings{lu2024scaffold,
  title={Scaffold-GS: Structured 3D Gaussians for View-Adaptive Rendering},
  author={Lu, Tao and Yu, Mulin and Xu, Linning and Xiangli, Yuanbo and Wang, Limin and Lin, Dahua and Dai, Bo},
  booktitle={Proceedings of the IEEE/CVF Conference on Computer Vision and Pattern Recognition},
  year={2024}
}

@inproceedings{zhang20233dshape2vecset,
  title={3DShape2VecSet: A 3D Shape Representation for Neural Fields and Generative Diffusion Models},
  author={Zhang, Biao and Yan, Jiapeng and Wonka, Peter},
  booktitle={ACM SIGGRAPH 2023 Conference Proceedings},
  year={2023}
}

@article{hong2024lrm,
  title={LRM: Large Reconstruction Model for Single Image to 3D},
  author={Hong, Yicong and Zhang, Kai and Gu, Jiuxiang and Bi, Sai and Zhou, Yang and Liu, Difan and Liu, Feng and Sunkavalli, Kalyan and Bui, Trung and Tan, Hao},
  journal={arXiv preprint arXiv:2311.04400},
  year={2024}
}

@inproceedings{wang2022detr3d,
  title={{DETR3D}: 3D Object Detection from Multi-view Images via 3D-to-2D Queries},
  author={Wang, Yue and Guizilini, Vitor and Zhang, Tianyuan and Wang, Yilun and Zhao, Hang and Solomon, Justin},
  booktitle={Conference on Robot Learning},
  year={2022},
  url={https://arxiv.org/abs/2110.06922}
}

@inproceedings{gao2024pointdetr3d,
  title={Point-DETR3D: Leveraging Imagery Data with Spatial Point Prior for Weakly Semi-supervised 3D Object Detection},
  author={Gao, Hongzhi and Chen, Zheng and Chen, Zehui and Chen, Lin and Liu, Jiaming and Zhang, Shanghang and Zhao, Feng},
  booktitle={Proceedings of the AAAI Conference on Artificial Intelligence},
  year={2024},
  url={https://arxiv.org/abs/2403.15317}
}

@inproceedings{zhang2024gslrm,
  title={{GS-LRM}: Large Reconstruction Model for 3D Gaussian Splatting},
  author={Zhang, Kai and Bi, Sai and Tan, Hao and Xiangli, Yuanbo and Zhao, Nanxuan and Sunkavalli, Kalyan and Xu, Zexiang},
  booktitle={European Conference on Computer Vision},
  year={2024},
  url={https://arxiv.org/abs/2404.19702}
}

@article{tang2024lgm,
  title={{LGM}: Large Multi-View Gaussian Model for High-Resolution 3D Content Creation},
  author={Tang, Jiaxiang and Chen, Zhaoxi and Chen, Xiaokang and Wang, Tengfei and Zeng, Gang and Liu, Ziwei},
  journal={arXiv preprint arXiv:2402.05054},
  year={2024},
  url={https://arxiv.org/abs/2402.05054}
}

@article{jin2024lvsm,
  title={{LVSM}: A Large View Synthesis Model with Minimal 3D Inductive Bias},
  author={Jin, Haian and Jiang, Hanwen and Tan, Hao and Zhang, Kai and Bi, Sai and Zhang, Tianyuan and Luan, Fujun and Snavely, Noah and Xu, Zexiang},
  journal={arXiv preprint arXiv:2410.17242},
  year={2024},
  url={https://arxiv.org/abs/2410.17242}
}

@article{asim2026scenetok,
  title={SceneTok: A Compressed, Diffusable Token Space for 3D Scenes},
  author={Asim, Mohammad and Wewer, Christopher and Lenssen, Jan Eric},
  journal={arXiv preprint arXiv:2602.18882},
  year={2026},
  url={https://arxiv.org/abs/2602.18882}
}

@article{zhang2026anchorsplat,
  title={AnchorSplat: Feed-Forward 3D Gaussian Splatting With 3D Geometric Priors},
  author={Zhang, Xiaoxue and Zheng, Xiaoxu and Yin, Yixuan and Zhao, Tiao and Tang, Kaihua and Mi, Michael Bi and Xu, Zhan and Chen, Dave Zhenyu},
  journal={arXiv preprint arXiv:2604.07053},
  year={2026},
  url={https://arxiv.org/abs/2604.07053}
}
